\documentclass{article}
\usepackage{iclr2026_conference,times}

\usepackage{amsmath,amsfonts,bm}

\def\eqref#1{equation~\ref{#1}}

\def\1{\bm{1}}

\DeclareMathAlphabet{\mathsfit}{\encodingdefault}{\sfdefault}{m}{sl}
\SetMathAlphabet{\mathsfit}{bold}{\encodingdefault}{\sfdefault}{bx}{n}

\usepackage{hyperref}
\usepackage{url}
\usepackage{booktabs}
\usepackage{multirow}
\usepackage{graphicx}
\usepackage{subcaption}
\usepackage{float}        
\usepackage{placeins}     

\iclrfinalcopy  

\begin{document}

\title{Reasoning or Rhetoric? An Empirical Analysis of Moral Reasoning Explanations in Large Language Models}

\author{%
  Aryan Kasat$^{*\dagger}$  \\
  TCS AI Practice
  \And
  Smriti Singh$^{*\dagger}$ \\
  UT Austin
  \And
  Aman Chadha$^{\dagger}$ \\
  Amazon GenAI
  \And
  Vinija Jain$^{\dagger}$ \\
  Meta GenAI
}

\maketitle

\fancyhead{}

\renewcommand{\thefootnote}{\fnsymbol{footnote}}

\footnotetext[1]{\textbf{Equal contribution.}}
\footnotetext[2]{\textbf{Work done outside role at the authors’ respective companies.}}

\renewcommand{\thefootnote}{\arabic{footnote}}

\begin{abstract}

Do large language models reason morally, or do they merely sound like they do? We investigate whether LLM responses to moral dilemmas exhibit genuine developmental progression through Kohlberg's stages of moral development, or whether alignment training instead produces reasoning-\emph{like} outputs that superficially resemble mature moral judgment without the underlying developmental trajectory. Using an LLM-as-judge scoring pipeline validated across three judge models, we classify more than 600 responses from 13 LLMs spanning a range of architectures, parameter scales, and training regimes across six classical moral dilemmas, and conduct ten complementary analyses to characterize the nature and internal coherence of the resulting patterns.

Our results reveal a striking inversion: responses overwhelmingly correspond to post-conventional reasoning (Stages~5--6) regardless of model size, architecture, or prompting strategy, the effective inverse of human developmental norms, where Stage~4 dominates. Most strikingly, a subset of models exhibit \emph{moral decoupling}: systematic inconsistency between stated moral justification and action choice, a form of logical incoherence that persists across scale and prompting strategy and represents a direct reasoning consistency failure independent of rhetorical sophistication. Model scale carries a statistically significant but practically small effect ($F(2,229)=6.05$, $p=0.003$, $\eta^2=0.050$, $d=0.55$); training type has no significant independent main effect ($p=0.065$); and models exhibit near-robotic cross-dilemma consistency (ICC~$>$~0.90), producing logically indistinguishable responses across semantically distinct moral problems. We posit that these patterns constitute evidence for \emph{moral ventriloquism}: the acquisition, through alignment training, of the rhetorical conventions of mature moral reasoning without the underlying developmental trajectory those conventions are meant to represent.

\end{abstract}


\section{Introduction}

A central question in AI alignment is not just whether large language models (LLMs) produce morally acceptable outputs, but how the moral reasoning explanations they generate relate to the processes that produce those outputs. Modern LLMs frequently generate detailed, sophisticated-sounding explanations when confronted with moral dilemmas, invoking abstract principles such as human dignity, social contracts, and universal rights. These behaviors are often interpreted as evidence that models possess genuine moral reasoning capabilities. At the same time, a growing body of research has raised serious concerns about interpreting model outputs as evidence of genuine underlying processes. Several studies suggest that LLMs may produce convincing reasoning patterns through statistical pattern completion rather than through explicit reasoning mechanisms \citep{turpin2023language, chen2025reasoning}. In particular, recent work questions whether chain-of-thought explanations correspond to the internal computational processes that generate model predictions, or whether intermediate tokens instead reflect learned stylistic patterns associated with reasoning discourse \citep{kambhampati2024planllms}. These concerns raise an important challenge: if models can produce explanations that \emph{resemble} reasoning without performing the underlying reasoning processes, behavioral evaluations based solely on model outputs may provide a systematically misleading picture of model capability.

Moral dilemmas are a particularly informative setting because they reliably elicit structured, principle-grounded reasoning explanations. Crucially, if LLMs genuinely develop moral reasoning capabilities as a function of scale and training, we would expect their outputs to mirror the developmental trajectory observed in humans, varying across contexts, progressing with scale, and converging toward the Stage~4 distribution that characterizes most human adults, not clustering uniformly at the highest stages regardless of model size or training procedure.

\begin{figure}[h]
  \centering
  \begin{subfigure}[t]{0.48\textwidth}
    \centering
    \includegraphics[width=\textwidth]{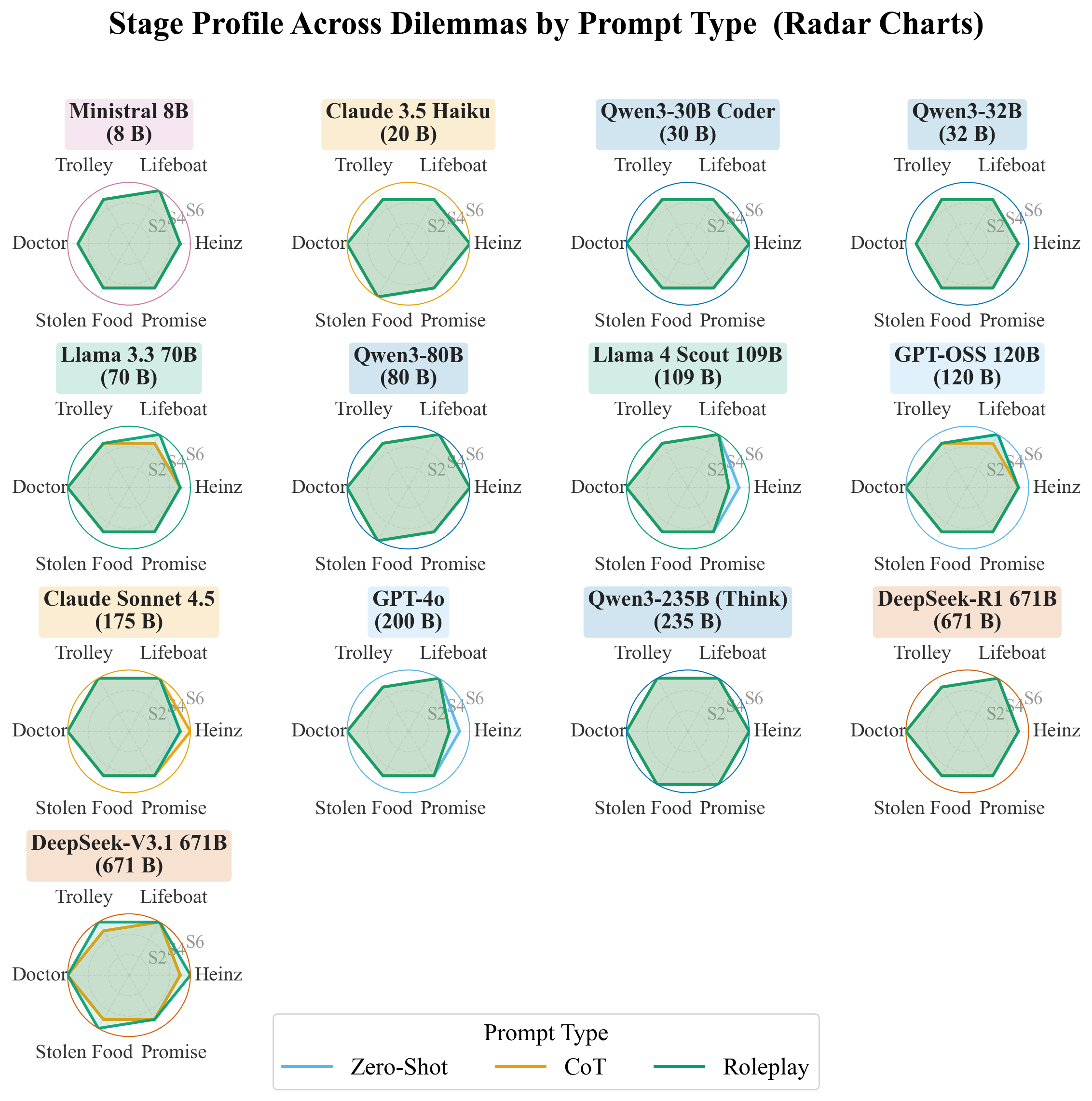}
    \caption{Stage profile across dilemmas by prompt type (radar charts). Each model's hexagonal profile is nearly identical across all six dilemmas and all three prompt types, illustrating near-zero cross-dilemma and cross-prompt variability.}
  \end{subfigure}\hfill
  \begin{subfigure}[t]{0.48\textwidth}
    \centering
    \includegraphics[width=\textwidth]{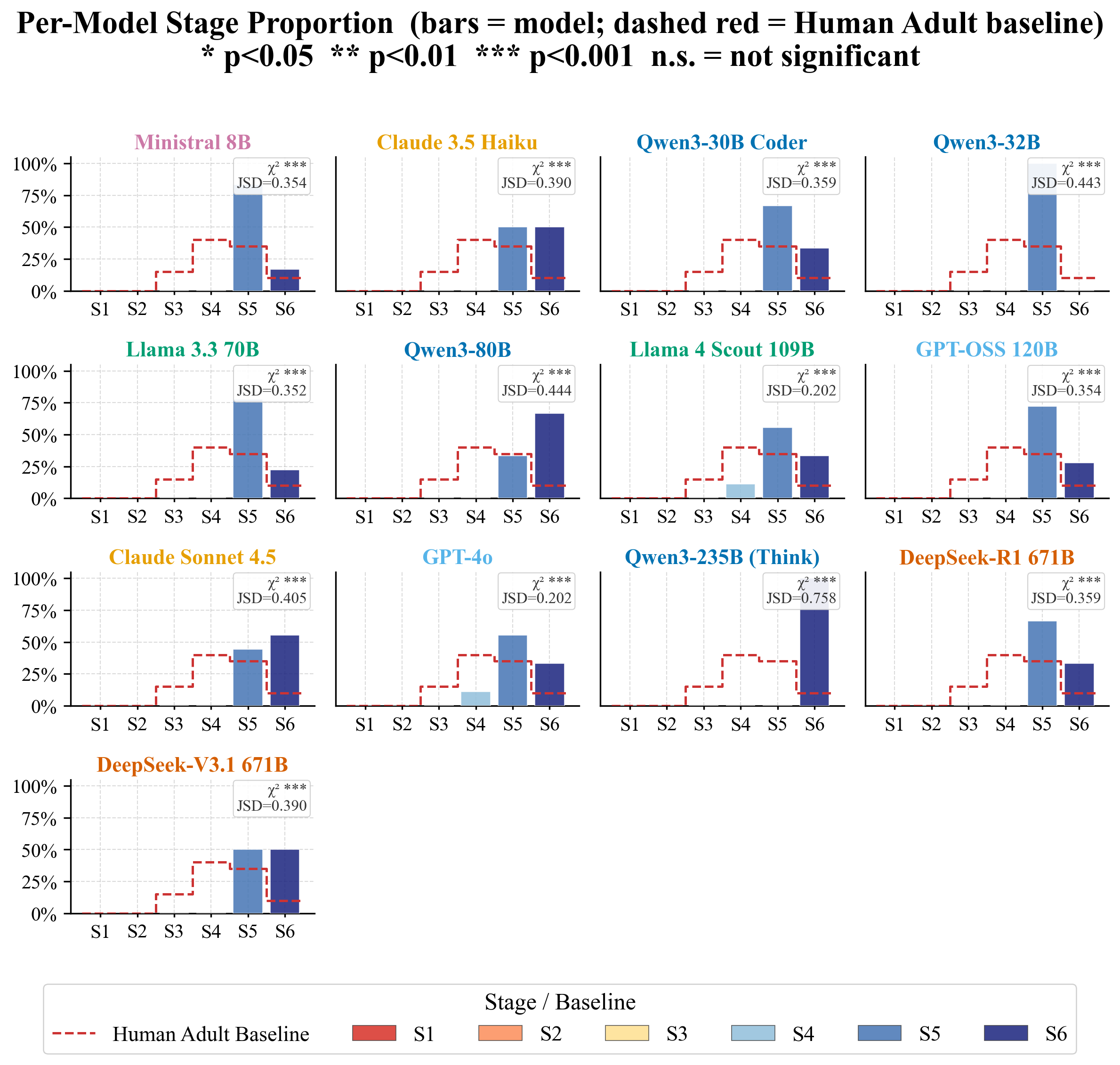}
    \caption{Per-model stage proportion vs.\ human adult baseline (dashed red). All 13 LLMs concentrate responses in Stages~5--6; Jensen--Shannon divergences from the human baseline are large ($\mathrm{JSD}\approx0.20$--$0.76$, all $\chi^2\!: p{<}0.001$), confirming the distributional inversion.}
  \end{subfigure}
  \caption{\textbf{Two Headline Anomalies.} \emph{Left:} Nearly identical radar profiles across all six morally distinct dilemmas and three prompt types reveal a cross-dilemma rigidity inconsistent with context-sensitive human moral reasoning. \emph{Right:} Per-model stage distributions diverge sharply from the human adult baseline --- LLMs produce the inverse of the Stage~4-dominant human pattern, clustering uniformly in post-conventional Stages~5--6. Together these patterns motivate the moral ventriloquism hypothesis.}
  \label{fig:intro_preview}
\end{figure}

To investigate this question, we conduct a large-scale empirical study evaluating moral reasoning patterns across 13 state-of-the-art LLMs, conducting ten quantitative analyses to characterize the nature, robustness, and internal coherence of these patterns. Figure~\ref{fig:intro_preview} previews two headline findings that shape our investigation. Our methodology is motivated by the following research questions:

\begin{description}
    \item[\textbf{RQ1}] Does model scale predict higher-stage moral reasoning explanations?
    \item[\textbf{RQ2}] Does prompting strategy systematically influence moral stage?
    \item[\textbf{RQ3}] Are model stage assignments consistent across dilemmas, and how does this compare to human moral variability?
    \item[\textbf{RQ4}] Do LLM stage distributions resemble human developmental norms?
    \item[\textbf{RQ5}] Do models practice what they preach: does stated moral reasoning correspond to the action choices models produce?
    \item[\textbf{RQ6}] Does scale affect moral reasoning independent of training type, or does training dominate once scale is controlled?
\end{description}

Our results reveal a striking inversion: across nearly all evaluated models, responses overwhelmingly correspond to Kohlberg's post-conventional stages (Stages~5--6), the effective inverse of the Stage~4-dominant human distribution, with remarkably little variation across model size, architecture, or prompting strategy. A subset of models additionally exhibit moral decoupling: producing high-stage justifications for low-stage action choices, and all models show near-robotic cross-dilemma consistency that is itself anomalous relative to human moral cognition.

Our primary contribution is the hypothesis of \emph{moral ventriloquism}: the acquisition, through alignment training, of the rhetorical conventions of mature moral reasoning without the underlying developmental trajectory. We use Kohlberg's framework as methodological scaffolding, exploiting its well-characterized human distribution as a diagnostic baseline rather than making claims about moral development per se. We provide convergent empirical evidence through ten complementary analyses spanning distributional comparisons, cross-dilemma consistency, action-reasoning alignment, linguistic profiling, and factorial decomposition of scale and training effects.


\section{Related Work}

\subsection{Reasoning Faithfulness and Moral Evaluation in LLMs}

Chain-of-thought prompting \citep{wei2022chain} generated widespread interest in using reasoning traces as a window into model cognition, but evidence quickly accumulated that this window may be opaque. \citet{turpin2023language} demonstrated that CoT explanations can systematically misrepresent the actual causes of model predictions, producing fluent justifications that do not correspond to the features that drove the prediction. \citet{chen2025reasoning} extended this finding to state-of-the-art reasoning models, showing that even Claude~3.7 Sonnet and DeepSeek~R1 routinely generate reasoning traces that do not reflect their actual inference process. \citet{kambhampati2024planllms} argue more fundamentally that auto-regressive LLMs function as non-veridical memory systems, with chain-of-thought traces better understood as learned stylistic registers than faithful records of deliberation. Together, these results motivate our core concern: if moral reasoning justifications can be fluent without being causally connected to model outputs, then stage-based evaluation of those justifications may not measure what it purports to measure.

On the evaluation side, \citet{awad2018moral} established the large-scale human baseline for moral preferences we use as our reference distribution. \citet{scherrer2023evaluating} treat LLM moral responses as direct evidence of ``beliefs encoded in LLMs'', an interpretation our results complicate, since post-conventional language may be a surface property of alignment rather than a reflection of underlying beliefs. \citet{ji2025moralbench} find substantial variation across moral dimensions and models, pointing to an important distinction our work sharpens: models may differ in \emph{which} principles they invoke while uniformly deploying the post-conventional register. \citet{jiao2025llmethics} measure whether models produce the \emph{right} moral outputs, but not whether those outputs reflect an underlying reasoning trajectory: the gap our Kohlberg-based diagnostic is designed to fill. \citet{zhou2024rethinking} show that grounding prompts in explicit moral theories improves classification accuracy; our Analysis~2 reaches a contrasting conclusion: even theory-invoking roleplay prompts do not significantly alter the stage distribution (Friedman $p=0.15$). \citet{garcia2024moral} found that human evaluators preferred LLM moral judgments and were only modestly better than chance at identifying AI-generated responses: exactly the surface indistinguishability that moral ventriloquism predicts. \citet{kudina2025structured} interpret structured prompting as ``unlocking'' latent moral competence; our results challenge this: scaffolding may improve benchmark accuracy while leaving the stage distribution unchanged.

\subsection{Alignment Training, Mechanistic Analysis, and Interventions}

RLHF \citep{ouyang2022training} and Constitutional AI \citep{bai2022constitutional} optimize for outputs judged ethically appropriate by human raters or AI proxies. This creates a systematic incentive: outputs invoking human rights, universal principles, and harm minimization receive high rewards, producing a strong learned association between moral dilemma contexts and Stage~5--6 rhetorical patterns independent of whether the model has internalized those principles. \citet{mahajan2025mapping} found that LLMs lack a transparent decision architecture for resolving safety principle conflicts, reinforcing our ventriloquism interpretation. \citet{schacht2025circuits} identified specialized neuron clusters for moral foundation dimensions via ablation studies; crucially, their circuit-level analysis does not address developmental-stage sensitivity, which our work targets. \citet{an2025moralreason} demonstrate that explicit RL training over structured moral reasoning data can improve generalization beyond the training distribution, a complement to our work: we show standard alignment produces the rhetoric without the substance; they show a qualitatively different training objective is required to begin closing that gap.

\section{Methodology}

Our goal is not to determine whether LLMs possess genuine moral reasoning capabilities
(a question that behavioral evaluation alone cannot answer), but to analyze the reasoning explanations models produce in moral dilemmas and the internal coherence of those explanations across contexts.

\subsection{Moral Reasoning Framework}

\textbf{Kohlberg's Theory of Moral Development.} We use Kohlberg's framework as methodological scaffolding rather than a claim about the ground truth of morality. Its six-stage structure (pre-conventional Stages~1--2: punishment avoidance and self-interest; conventional Stages~3--4: social expectations and rule-following; post-conventional Stages~5--6: abstract principles and universal ethics) provides a \emph{distributional diagnostic}: its well-characterized human distribution (Stage~4-dominant, with Stage~6 rare) gives a principled empirical baseline against which model outputs can be compared. Systematic deviation from this baseline is the diagnostic signal; no prior work has exploited this property to probe LLM alignment at scale.

\subsection{Automated Scoring Pipeline}

\textbf{LLM-as-Judge Classification.} Responses are scored using an LLM-as-judge system in which a scoring model is presented with each model response and prompted to classify it according to Kohlberg's framework. For each response, the scoring model outputs: (1) a primary Kohlberg stage assignment, (2) a confidence score for the assignment, and (3) a natural language explanation justifying the classification. To assess reliability, we evaluate stage classifications across three judge models (GPT-4, Claude Sonnet, and Llama-3) and report inter-judge agreement (see Appendix~\ref{app:analysis7}). The pipeline also records a secondary stage assignment for borderline responses, enabling analysis of distributional uncertainty.

\subsection{Dilemmas and Prompt Configurations}

\textbf{Moral Dilemmas.} We evaluate models on six classical moral dilemmas drawn from the moral psychology literature: the Heinz dilemma, the trolley problem, the lifeboat dilemma, a doctor truth-telling dilemma, a stolen food dilemma, and a broken promise dilemma. These dilemmas were selected to span a range of moral dimensions including harm, fairness, property, authority, and loyalty.

\textbf{Prompt Configurations.} Each dilemma is presented under three prompt configurations: (1) zero-shot prompting, in which the model is asked directly for its response; (2) chain-of-thought prompting, in which the model is explicitly instructed to reason step-by-step before answering; and (3) roleplay prompting, in which the model is asked to respond as a moral philosopher. This design allows us to assess whether prompting strategy systematically influences the stage of moral reasoning produced.

\subsection{Analytical Framework}

Our ten analyses are organized around a single diagnostic logic: if LLMs genuinely develop moral reasoning, their outputs should vary with scale, respond to prompting, shift across dilemma contexts, and converge toward human developmental distributions. Departures from each of these expectations (individually suggestive, collectively definitive) constitute the evidential basis for the moral ventriloquism hypothesis.

We first establish the \textbf{scale--stage relationship} (Analysis~1) and \textbf{prompting sensitivity} (Analysis~2) as baseline tests of whether model capability and elicitation strategy can produce developmental differentiation. We then examine \textbf{cross-dilemma consistency} (Analysis~3) as a direct test of logical rigidity: whether models produce indistinguishable responses across semantically distinct problems. \textbf{Distributional comparison against human norms} (Analysis~4) provides the sharpest population-level test of the genuine development hypothesis. \textbf{Action--reasoning alignment} (Analysis~5) is the core logical coherence test: whether stated justifications are consistent with actual choices. \textbf{Linguistic profiling} (Analysis~6) identifies the training-regime fingerprints in moral vocabulary, isolating RLHF as the primary rhetorical driver. Finally, \textbf{factorial decomposition} (Analysis~8) disentangles the independent contributions of scale and training type under controlled conditions.

Analyses~7, 9, and~10 (response quality, sub-capability thresholds, and stage transition dynamics) provide corroborating mechanistic detail and are reported in full in Appendix~\ref{app:analysis7}--\ref{app:analysis10}. Technical specifications for all analyses are in Appendix~\ref{app:stats}.

\section{Experimental Setup}

\subsection{Models}

We evaluate 13 LLMs spanning a range of architectures, parameter scales, and training regimes, including both frontier and open-source models. The evaluated models include Qwen3-235B (Thinking), Qwen3-80B, Claude~Sonnet~4.5, Claude~3.5~Haiku, DeepSeek~V3.1, DeepSeek~R1, Qwen3-30B Coder, GPT-OSS-120B, GPT-4o, Llama~4~Scout, Llama~3.3~70B, Ministral~8B, and Qwen3-32B. For Analysis~8, models are grouped into three scale tiers (Small:~8--32B; Mid:~70--120B; Large:~175--671B) and three training-type categories (Base-RLHF, Coding-Tuned, Reasoning-Tuned), yielding a $3\times3$ factorial design over 234 observations.

\subsection{Dataset Statistics}

Our evaluation covers 13 models $\times$ 6 dilemmas $\times$ 3 prompt configurations $\times$ 3 responses per configuration, yielding $>$600 total responses and 234 observations for the factorial ANOVA (see Appendix~\ref{app:stats} for full dataset statistics).

\section{Results}
We evaluate patterns in the moral reasoning explanations produced by 13 modern LLMs. Responses are classified into Kohlberg stages using the automated scoring pipeline described in Section~3. We report results from ten complementary analyses below; Analyses~7, 9, and~10 are reported in full in Appendix~\ref{app:analysis7}--\ref{app:analysis10}.

\subsection{Scale and Moral Stage}

\textbf{RQ1.} Spearman rank correlation between log-parameter count and mean moral stage yields a moderate positive relationship ($\rho = 0.52$, $p < 0.05$): bigger models generally score higher. However, this trend exhibits clear diminishing returns past approximately 70B parameters, and mean stages across the entire model set span less than one full stage point: from 5.00 (Qwen3-32B) to 6.00 (Qwen3-235B Thinking). Even the smallest evaluated model (Ministral~8B, mean stage~5.17) produces predominantly post-conventional reasoning. Per-model stage means are in Appendix~\ref{app:model_table}; the scale--stage relationship is shown in Figure~\ref{fig:intro_preview} (left).

\subsection{Prompting Strategy Effects}

\textbf{RQ2.} A repeated-measures Friedman test across the three prompt configurations yields no statistically significant difference in moral stage ($\chi^2(2) = 3.84$, $p = 0.15$), with no prompt pair differing significantly after Bonferroni correction. The gap between zero-shot (mean~5.20) and roleplay (mean~5.61) spans less than half a stage point; all configurations produce predominantly Stage~5--6 outputs (full table and figure in Appendix~\ref{app:analysis7}).

\textbf{Finding:} Prompting strategy has no significant effect on fundamental moral stage. Post-conventional reasoning is baked in, not prompted out.

\subsection{Cross-Dilemma Consistency}

\textbf{RQ3.} Intraclass Correlation Coefficients computed per model across the six dilemmas reveal that models produce logically indistinguishable responses regardless of the dilemma presented (ICC~$>$~0.90 for all evaluated models; see Figure~\ref{fig:intro_preview}, right). This rigidity is anomalous relative to human moral cognition: empirical studies consistently find that individuals reason at different stages depending on the dilemma domain, shifting between conventional and post-conventional reasoning as a function of perceived stakes, familiarity, and cultural context.

The highest mean stage (trolley problem, 5.61) and the lowest (stolen food dilemma, 5.28) differ by only 0.33 stage points (full dilemma table in Appendix~\ref{app:analysis7}). LLMs are not adapting their reasoning to the problem; they are producing a fixed rhetorical register regardless of the stimulus. In the logical reasoning sense, a system that produces identical outputs to non-identical inputs is not reasoning about those inputs at all: the output is a function of training, not of the problem.

\subsection{Comparison with Human Developmental Distributions}

\textbf{RQ4.} Chi-squared goodness-of-fit tests comparing each model's stage distribution against empirical human developmental norms reject the null hypothesis of distributional equivalence for all evaluated models ($p < 0.001$ in all cases). Jensen-Shannon divergence between model distributions and the human reference distribution is uniformly high (mean JS~$= 0.71$), confirming that model outputs do not approximate human developmental distributions.

The LLM distributions we observe are effectively the inverse of the human pattern: Stages~5--6 account for 86\% of all model responses, while Stage~4 accounts for only 10\% and Stages~1--3 together for just 4\%. Table~\ref{tab:stage_distribution} presents the full distribution. Notably, a small number of heavily RLHF-tuned frontier models show distributions marginally closer to human norms, driven by somewhat higher Stage~4 representation, a result we examine further in Analysis~8.

\begin{table}[t]
\centering
\caption{Distribution of Kohlberg stage classifications across all model responses ($>$600 responses). Chi-squared goodness-of-fit against human developmental norms: $p < 0.001$. Mean Jensen-Shannon divergence from human reference: 0.71.}
\label{tab:stage_distribution}
\begin{tabular}{lcc}
\toprule
Stage & \% of LLM Responses & Typical Human \% \\
\midrule
Stage 1 & 0\%  & $\sim$5\%  \\
Stage 2 & 1\%  & $\sim$10\% \\
Stage 3 & 3\%  & $\sim$15\% \\
Stage 4 & 10\% & $\sim$50\% \\
Stage 5 & 55\% & $\sim$15\% \\
Stage 6 & 31\% & $\sim$5\%  \\
\bottomrule
\end{tabular}
\end{table}

\begin{figure}[t]
  \centering
  \includegraphics[width=0.72\textwidth]{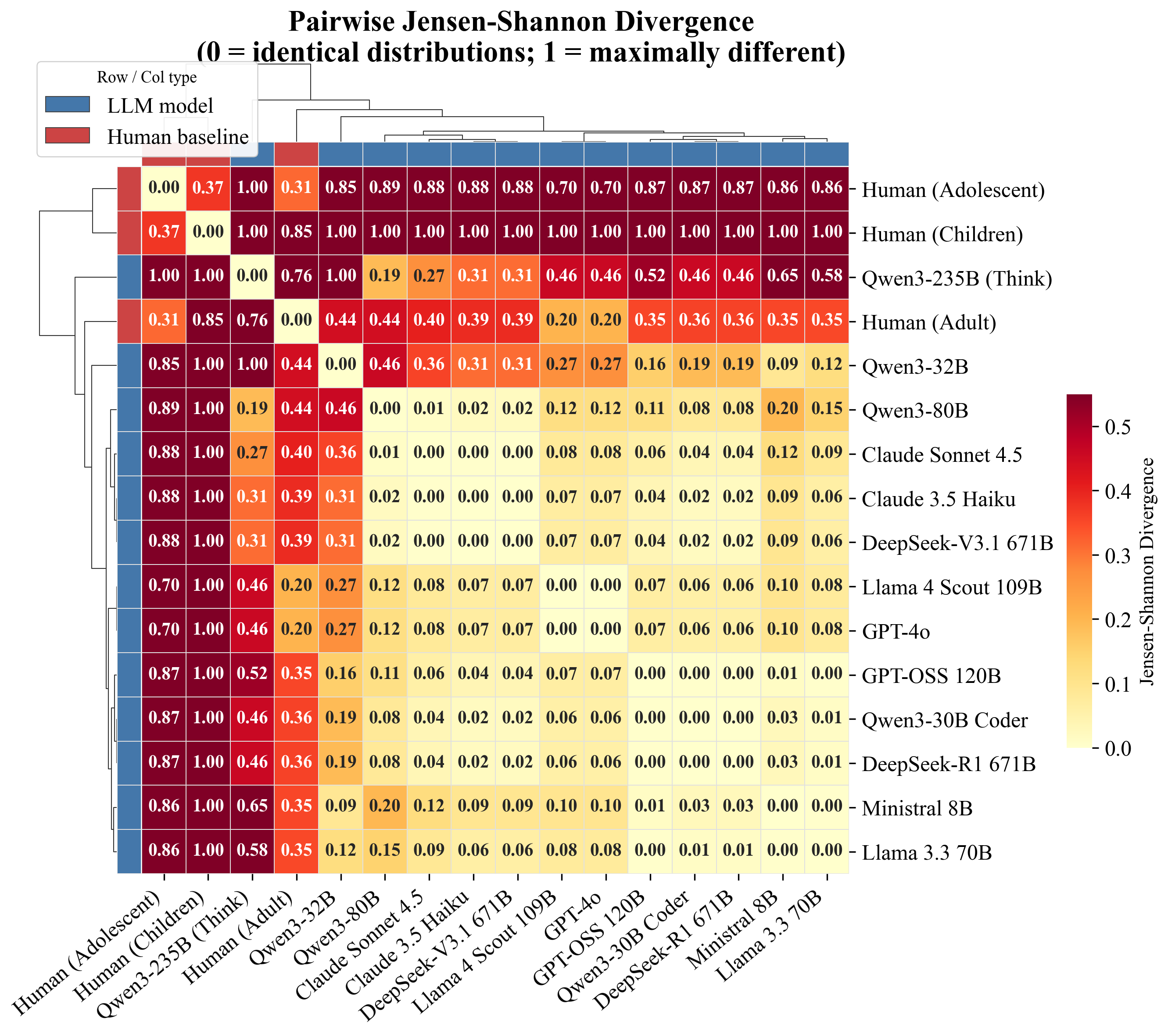}
  \caption{\textbf{RQ4: Divergence from Human Norms.} Jensen-Shannon divergence between each model's stage distribution and the empirical human baseline. All models yield JS~$>$~0.60 (mean~0.71), confirming that model outputs do not approximate human developmental distributions regardless of scale or training type.}
  \label{fig:analysis4}
\end{figure}

\subsection{Action--Reasoning Alignment and Moral Decoupling}

\textbf{RQ5.} Action--reasoning cross-tabulation and Cram\'{e}r's~$V$ analysis reveals strong statistical association overall ($V = 0.61$, $p < 0.001$), indicating that for most models, stated reasoning and action choices are broadly aligned. However, this aggregate finding conceals a critical heterogeneity.

A subset of models exhibit \emph{moral decoupling}: they consistently produce high-stage vocabulary and argumentation (Stage~5--6) while selecting action choices more consistent with lower-stage reasoning (Stage~3--4). This is a logical incoherence failure: the model's stated justification and its actual choice operate on different tracks. The pattern is most pronounced in mid-tier models (GPT-OSS-120B, Llama~4~Scout, Qwen3-80B show the largest gaps) and least pronounced in the largest reasoning-tuned models (DeepSeek~R1, Qwen3-235B Thinking), suggesting reasoning-focused training partially closes the gap but does not eliminate it. Figure~\ref{fig:analysis5} shows the per-model consistency scores; the cross-tabulation heatmap is in Appendix~\ref{app:figures}.

\begin{figure}[t]
  \centering
  \includegraphics[width=0.72\textwidth]{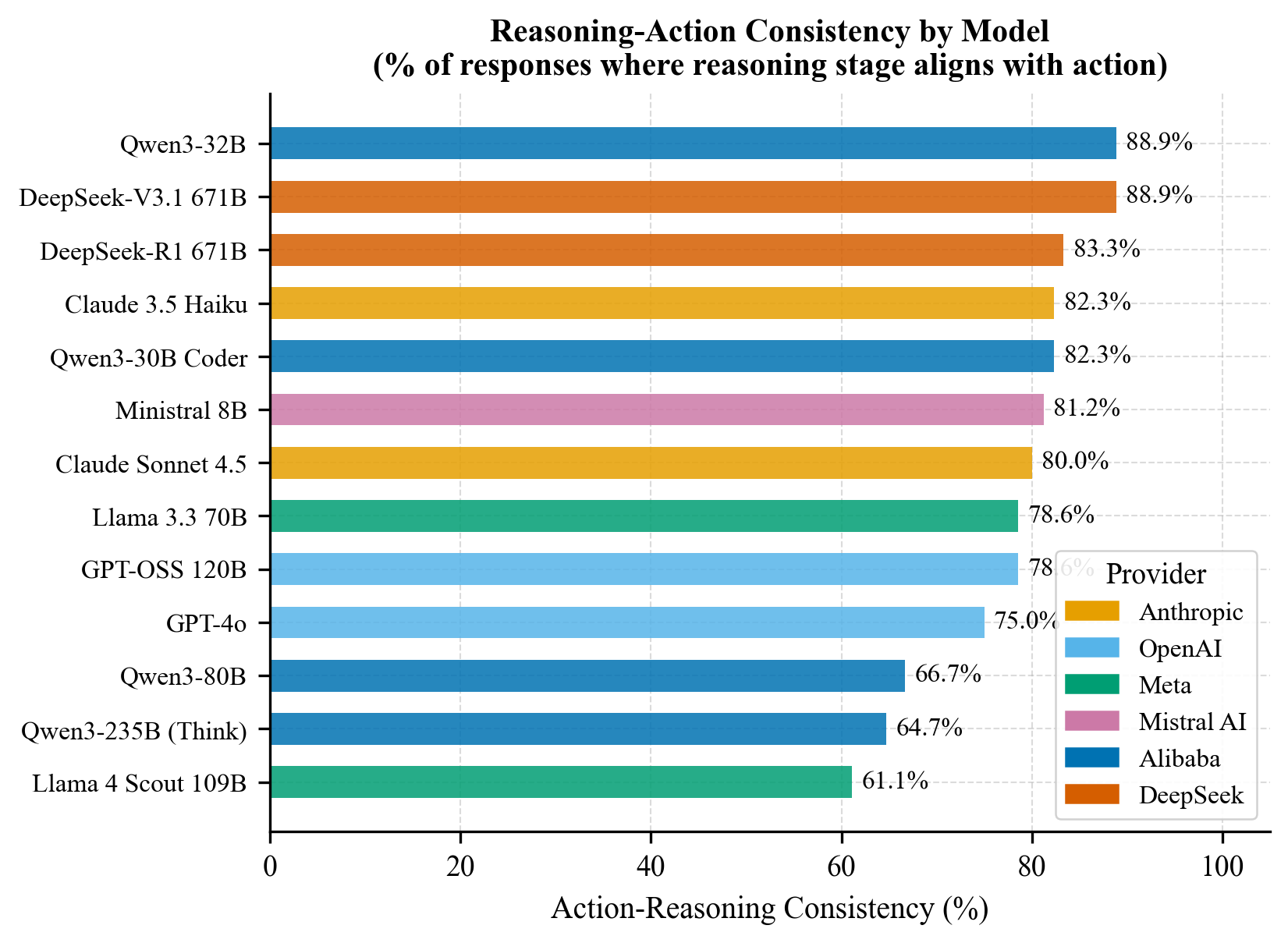}
  \caption{\textbf{RQ5: Moral Decoupling by Model.} Per-model action--reasoning consistency scores. Mid-tier models show the largest gap between stated justification stage and action choice; large reasoning-tuned models the smallest. This logical incoherence is the most direct behavioral signature of moral ventriloquism.}
  \label{fig:analysis5}
\end{figure}

\subsection{Linguistic Patterns Across Model Families}

TF-IDF keyword extraction and PCA dimensionality reduction reveal that model families occupy distinct regions of moral vocabulary space. RLHF-aligned models of all sizes employ substantially richer moral vocabulary than their base or coding-tuned counterparts, using a broader range of terms associated with rights, dignity, fairness, and contextual nuance. Reasoning-tuned models produce structurally more complex responses with more explicit hedging and principle enumeration, but their moral vocabulary, once normalized for length, overlaps substantially with that of general RLHF models.

This supports the interpretation that RLHF training is the primary mechanism driving post-conventional moral language acquisition, independent of scale: small RLHF-aligned models share a recognizable moral vocabulary profile with much larger models from the same family. Coding-tuned models (e.g., Qwen3-30B~Coder) produce noticeably sparser moral vocabulary, suggesting that rhetorical richness is a function of alignment procedure rather than raw capability.

\begin{figure}[t]
  \centering
  \begin{subfigure}[t]{0.52\textwidth}
    \centering
    \includegraphics[width=\textwidth]{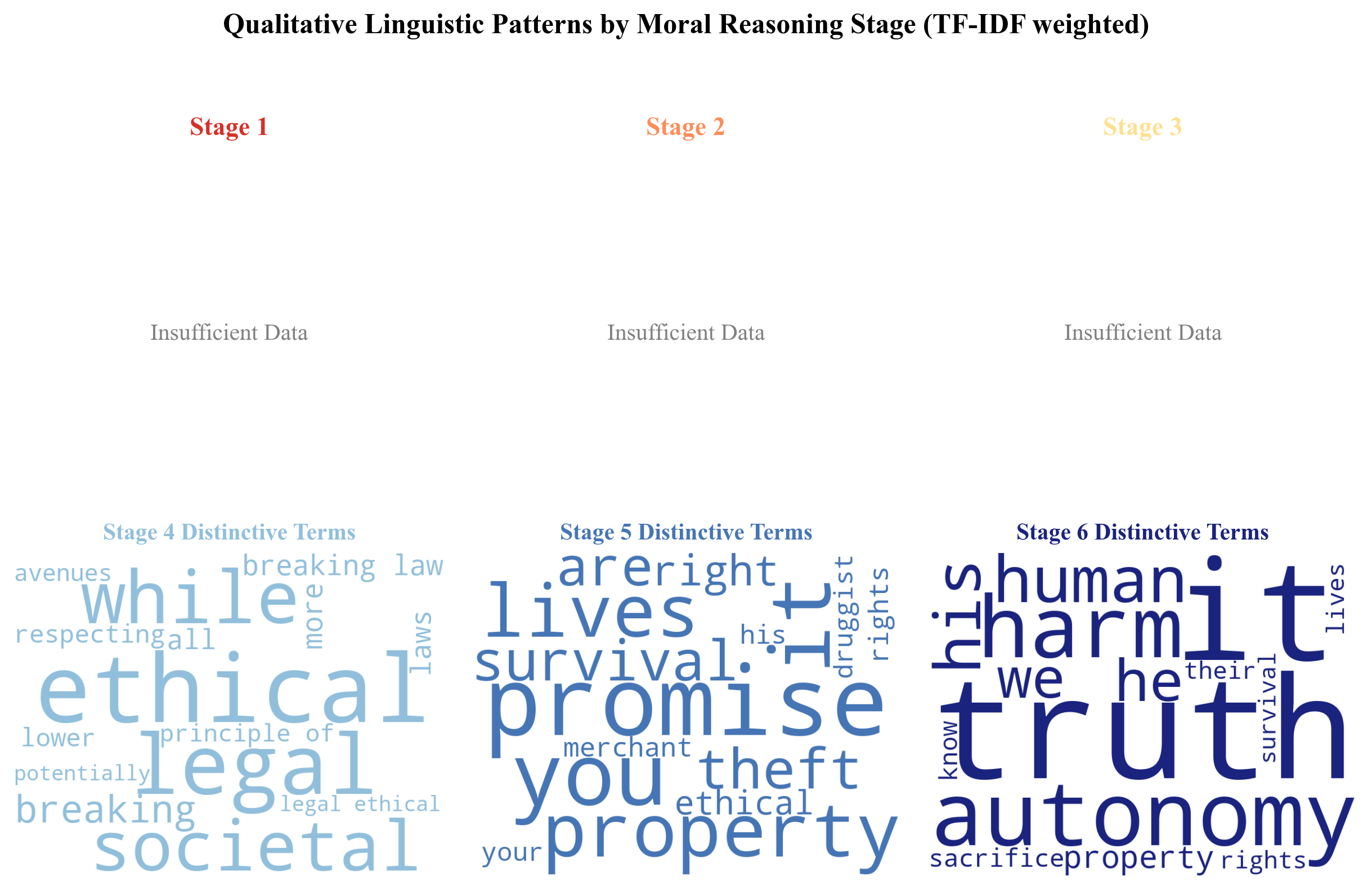}
    \caption{Moral vocabulary word clouds by Kohlberg stage. Post-conventional stages (5--6) are saturated with rights, dignity, and principle terms; pre-conventional stages are sparse.}
  \end{subfigure}\hfill
  \begin{subfigure}[t]{0.44\textwidth}
    \centering
    \includegraphics[width=\textwidth]{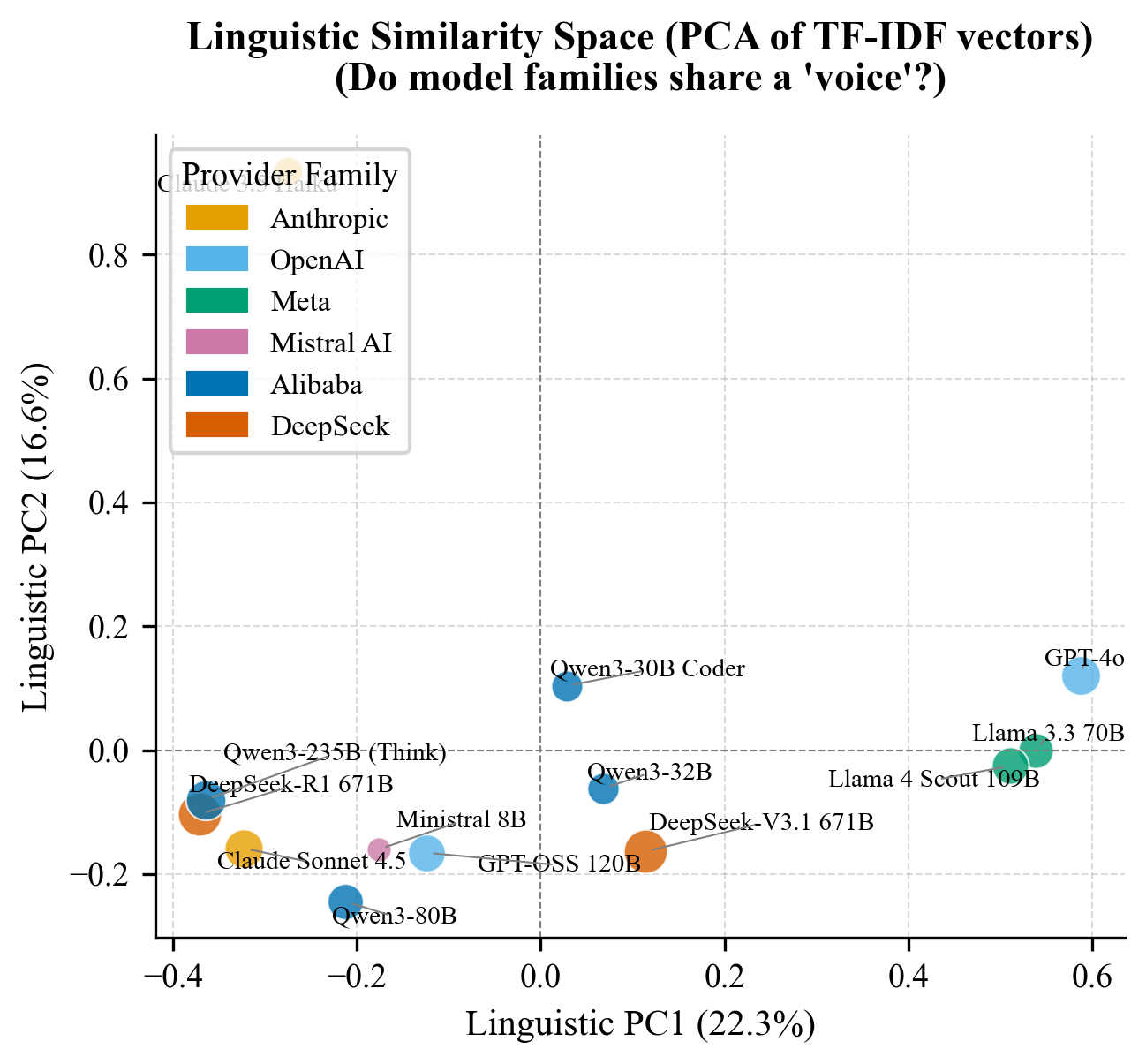}
    \caption{PCA of moral vocabulary. Model families cluster by training regime (RLHF vs.\ coding-tuned), not by scale: RLHF is the primary driver of rich post-conventional rhetoric.}
  \end{subfigure}
  \caption{\textbf{Analysis~6: Linguistic Patterns.} Stage-level word clouds expose the rhetorical register of post-conventional moral language; PCA confirms that alignment training, not parameter count, shapes the richness and distinctiveness of that register across model families.}
  \label{fig:analysis6}
\end{figure}

\subsection{Scale vs.\ Training Type: Factorial Decomposition}

\textbf{RQ6.} Factorial ANOVA over 234 observations (Scale Group $\times$ Training Type) finds scale is a statistically significant but practically small independent predictor ($F(2,229)=6.05$, $p=0.003$, $\eta^2=0.050$, $d=0.55$), confirmed by Kruskal-Wallis ($H=12.78$, $p=0.002$) and Welch ANOVA ($F=6.26$, $p=0.002$). Training Type has no significant main effect ($p=0.065$), though within the Large scale group, Reasoning-Tuned models score higher than Base-RLHF (Tukey, $p=0.039$). Full test table is in Appendix~\ref{app:anova}.

Scale shifts an already post-conventional floor modestly upward; mean stages never fall below 5.00 even in the Small group (8--32B). Neither scale nor training type produces the Stage~2-to-6 developmental arc Kohlberg's framework was designed to capture.

\begin{figure}[t]
  \centering
  \includegraphics[width=0.6\textwidth]{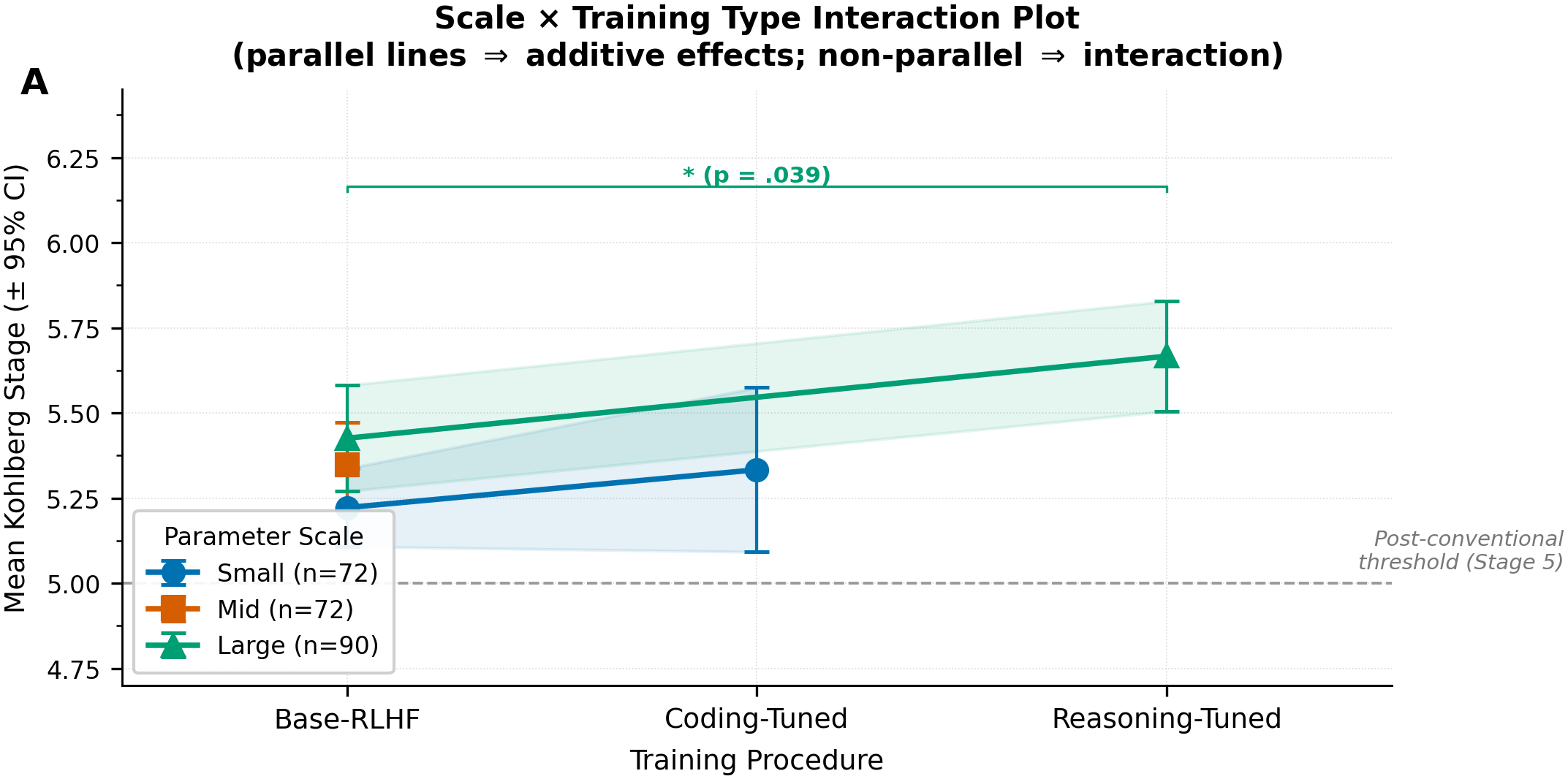}
  \caption{\textbf{RQ6: Factorial Decomposition.} Scale~$\times$~Training Type interaction. Scale reaches significance ($p=0.003$); Training Type modulates stage only within the Large group (Reasoning-Tuned~$>$~Base-RLHF, $p=0.039$). Stage score distributions by scale group in Appendix~\ref{app:analysis9}.}
  \label{fig:analysis8}
\end{figure}

\section{Discussion}

Our ten analyses converge on a coherent picture. We argue the results are best understood as \emph{moral ventriloquism}: the acquisition, through alignment training, of the rhetorical conventions of mature moral reasoning without the underlying cognitive architecture those conventions are meant to reflect. We use Kohlberg's developmental framework as diagnostic scaffolding throughout: our core claim is about logical coherence between justification and action, not about moral development per se.

We distinguish two claims our evidence supports to different degrees. Our \emph{empirical finding} that LLMs produce post-conventional moral \emph{language} is directly demonstrated by the stage distributions, ICC results, and linguistic analyses. Our \emph{interpretive hypothesis} that this language does not reflect genuine moral reasoning is supported behaviorally by the decoupling and distributional inversion results, but would require mechanistic evidence to establish with certainty.

\textbf{Decoupling and logical incoherence (RQ5).} The moral decoupling finding is the most direct contribution of this paper to the logical reasoning literature. A subset of models, concentrated in the mid-tier, consistently produce high-stage justifications for low-stage action choices: their stated reasoning and actual behavior are logically inconsistent. This pattern is related to, but distinct from, the sycophancy problem \citep{turpin2023language}. Sycophancy describes models that adjust their stated position in response to perceived social pressure from the user, producing outputs that track user preference rather than truth. Moral decoupling is a different failure mode: it is not elicited by external pressure but appears as a stable internal inconsistency, present across all three prompting strategies (RQ2) and across dilemmas (RQ3), regardless of whether any social pressure is applied. This suggests decoupling is a structural consequence of how alignment training installs rhetorical patterns independently of the decision process, rather than a context-sensitive sycophantic response. The distinction matters for mitigation: sycophancy may be addressed by prompting strategies that reduce social pressure cues, whereas moral decoupling likely requires training-level interventions that couple justification generation to action selection.

\textbf{Rigidity and distributional inversion (RQ3--4).} The hyper-consistency finding (ICC~$>$~0.90) is a logical rigidity result: LLMs produce responses that are statistically indistinguishable across six semantically distinct moral problems. This is not contextual robustness; it is the absence of sensitivity to problem structure that genuine reasoning would require. Framed in terms of the workshop's core theme, moral dilemmas are a domain where avoiding contradictions across related questions requires sensitivity to the specific structure of each problem: models that produce identical stage assignments regardless of whether a dilemma involves harm, property, or loyalty are contradicting the premise that they are reasoning at all. The distributional inversion (86\% Stages~5--6 vs.\ $\sim$50\% Stage~4 in humans; mean JS~$=0.71$) corroborates this: the stage distribution is set by training, not by reasoning about the specific dilemma.

\textbf{Scale and prompting (RQ1--2).} Scale is statistically significant but practically small ($\eta^2=0.050$, $d=0.55$), spanning less than one stage point; even the smallest models produce post-conventional outputs. Prompting strategy has no significant effect (Friedman $p=0.15$). Post-conventional moral language is a stable learned property, neither withheld by low capacity nor unlocked by careful elicitation \citep{kambhampati2024planllms}. Sub-capability analysis (Appendix~\ref{app:analysis9}) confirms that what scale contributes is primarily surface rhetorical richness: semantic density and syntactic complexity, rather than deeper moral cognition.

A natural concern is whether our LLM-as-judge pipeline is itself subject to the ventriloquism it detects. We address this in Section~\ref{sec:limitations}: uniform judge inflation cannot explain the ICC finding, inter-judge agreement is high across three architecturally distinct models, and the decoupling finding is a within-response comparison less susceptible to uniform inflation.

The practical implication is direct: behavioral evaluation of moral reasoning faces a fundamental identification problem. Models that have learned to produce post-conventional rhetoric will pass stage-based evaluations regardless of whether the underlying reasoning is genuine. Moving beyond this requires evaluation designs that probe action--reasoning coherence and contextual sensitivity, or mechanistic methods that examine internal representations rather than surface outputs.

\section{Limitations}
\label{sec:limitations}

Our behavioral evaluation cannot prove the mechanistic claim that RLHF \emph{produces} ventriloquism rather than correlating with it; establishing causality requires mechanistic interpretability methods beyond our current scope. A related concern is that Kohlberg's highest stages are definitionally similar to RLHF training objectives, making the ``rhetoric without substance'' interpretation potentially circular: models may score high simply because the scoring framework and the training signal share the same rhetorical target. Our coding-tuned models provide a partial control condition: Qwen3-30B Coder receives substantially less moral RLHF signal than its general-purpose counterparts and produces noticeably sparser moral vocabulary and lower mean stage scores despite comparable base capabilities, consistent with the interpretation that RLHF shapes the rhetorical output independently of underlying reasoning capacity. Additional triangulation comes from the reasoning-tuned models (DeepSeek~R1, Qwen3-235B Thinking): despite achieving the highest stage scores, their moral vocabulary profiles are distinct from RLHF-aligned models of similar scale, suggesting that reasoning-focused training and moral RLHF operate through separable mechanisms rather than a single confounded signal. Full mechanistic disambiguation remains an open problem. Our LLM-as-judge pipeline uses RLHF-aligned judges that may pattern-match to post-conventional rhetoric, though the ICC consistency and cross-architecture inter-judge agreement partially mitigate this (Appendix~\ref{app:analysis7}), and future work should validate against human annotations. Kohlberg's framework is contested in developmental psychology, but its well-characterized human distribution makes it uniquely suited as a distributional diagnostic: no prior work has exploited this property to probe LLM alignment at scale. Our 13-model sample is the largest Kohlberg-based evaluation to date; our six dilemmas span harm, property, authority, loyalty, and truth-telling dimensions, covering the domain range used in the human ICC studies we cite, though findings may not generalize to models trained on substantially different data mixtures.

\section{Conclusion}

We conduct an empirical study evaluating moral reasoning across 13 state-of-the-art LLMs, finding that models overwhelmingly produce post-conventional moral language (Stages~5--6) regardless of scale, architecture, or prompting strategy: the effective inverse of human developmental norms, with near-robotic cross-dilemma consistency and, in a subset of models, moral decoupling between stated justifications and action choices.

We posit that these findings constitute evidence for \emph{moral ventriloquism}: the acquisition, through alignment training, of the rhetorical conventions of mature moral reasoning without the underlying developmental trajectory (a correlational inference, not a proven causal claim).

These results have three concrete implications. First, moral stage as measured by output classification is not a reliable indicator of genuine moral reasoning capability or deployment readiness. Second, evaluating moral reasoning capability requires moving beyond output classification toward methods that probe action--reasoning coherence, contextual sensitivity, and internal representation fidelity. One concrete direction: adversarial dilemma pairs in which the rhetorically prestigious response (high-stage language) conflicts with the logically correct action, designed to directly measure whether a model tracks its stated justification or defaults to a training-shaped rhetorical pattern. Models exhibiting moral ventriloquism should show systematic divergence on such pairs; models with genuine action--justification coherence should not.

\bibliography{iclr2026_conference}
\bibliographystyle{iclr2026_conference}

\appendix

\section{Statistical Details}
\label{app:stats}

\subsection{Analysis 8: Factorial ANOVA Design}
\label{app:anova}

The two-way factorial ANOVA was conducted on 234 observations across 13 models grouped into Scale (Small:~8--32B, Mid:~70--120B, Large:~175--671B) and Training Type (Base-RLHF, Coding-Tuned, Reasoning-Tuned) categories. Sequential (Type~I) sums of squares were used with Scale entered first to partial out its contribution before assessing Training Type. Homogeneity of variance was assessed via Levene's test; Welch ANOVA and Kruskal-Wallis were used as convergent validators given unequal cell sizes. Post-hoc comparisons used Tukey's HSD for parametric and Bonferroni-corrected Mann-Whitney~U for non-parametric contrasts.

\begin{table}[h]
\centering
\caption{Factorial analysis results: Scale~$\times$~Training Type ANOVA on moral stage scores ($N=234$). Scale is a statistically significant independent predictor; Training Type shows a conditional interaction within the Large scale group only.}
\label{tab:anova_results}
\begin{tabular}{llccc}
\toprule
Test & Effect & Statistic & $p$-value & Effect Size \\
\midrule
Sequential ANOVA     & Scale         & $F(2,229)=6.05$ & 0.003** & $\eta^2=0.050$, $\omega^2=0.041$ \\
Sequential ANOVA     & Training Type & $F(2,229)=2.76$ & 0.065   & -- \\
Kruskal-Wallis       & Scale         & $H=12.78$       & 0.002** & -- \\
Welch ANOVA          & Scale         & $F=6.26$        & 0.002** & -- \\
Mann-Whitney U (Bonferroni) & Large vs.\ Small & --   & 0.001** & $d=0.55$ \\
\bottomrule
\multicolumn{5}{l}{\small{** $p<0.01$. Tukey post-hoc: Reasoning-Tuned $>$ Base-RLHF within Large group, $p=0.039$.}}
\end{tabular}
\end{table}

\subsection{Analysis 3: ICC Computation}

Intraclass Correlation Coefficients were computed using a two-way mixed-effects model (ICC(3,1)) with models as subjects and dilemmas as raters, following Shrout \& Fleiss conventions. ICC values above 0.90 are conventionally interpreted as excellent reliability; the application here to moral stage consistency means that dilemmas explain negligible variance in stage assignment relative to the model-level mean.

\subsection{Analysis 10: Entropy and Gini}
\label{app:analysis10}

Shannon entropy per model was computed over the six-category stage distribution ($H = -\sum_i p_i \log_2 p_i$, maximum~$= 2.58$ bits for a uniform distribution over six stages). Gini coefficients were computed over stage frequency distributions to measure concentration. Human reference entropy and Gini values were estimated from the developmental distributions reported in \citet{awad2018moral} and the broader moral development literature. Mean entropy across all scale groups was $H = 1.82$ bits (vs.\ $>$0.60 Gini typical of human stage-consolidated distributions, mean Gini~$= 0.31$ for LLMs), confirming non-sequential, unstable progression inconsistent with genuine developmental consolidation.

\begin{figure}[h]
  \centering
  \begin{subfigure}[t]{0.48\textwidth}
    \centering
    \includegraphics[width=\textwidth]{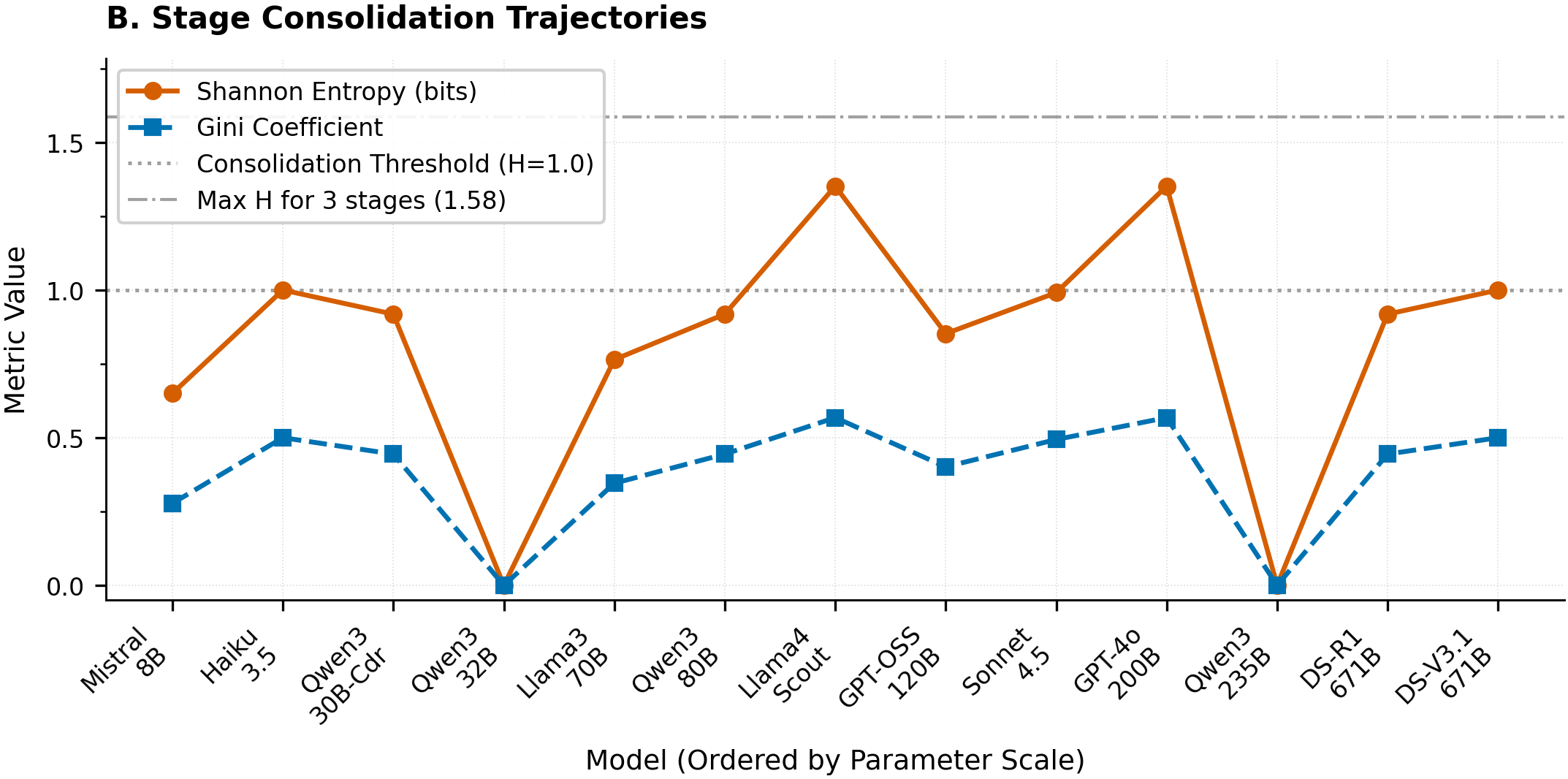}
    \caption{Shannon entropy trajectory across scale tiers (mean $H=1.82$ bits).}
  \end{subfigure}\hfill
  \begin{subfigure}[t]{0.48\textwidth}
    \centering
    \includegraphics[width=\textwidth]{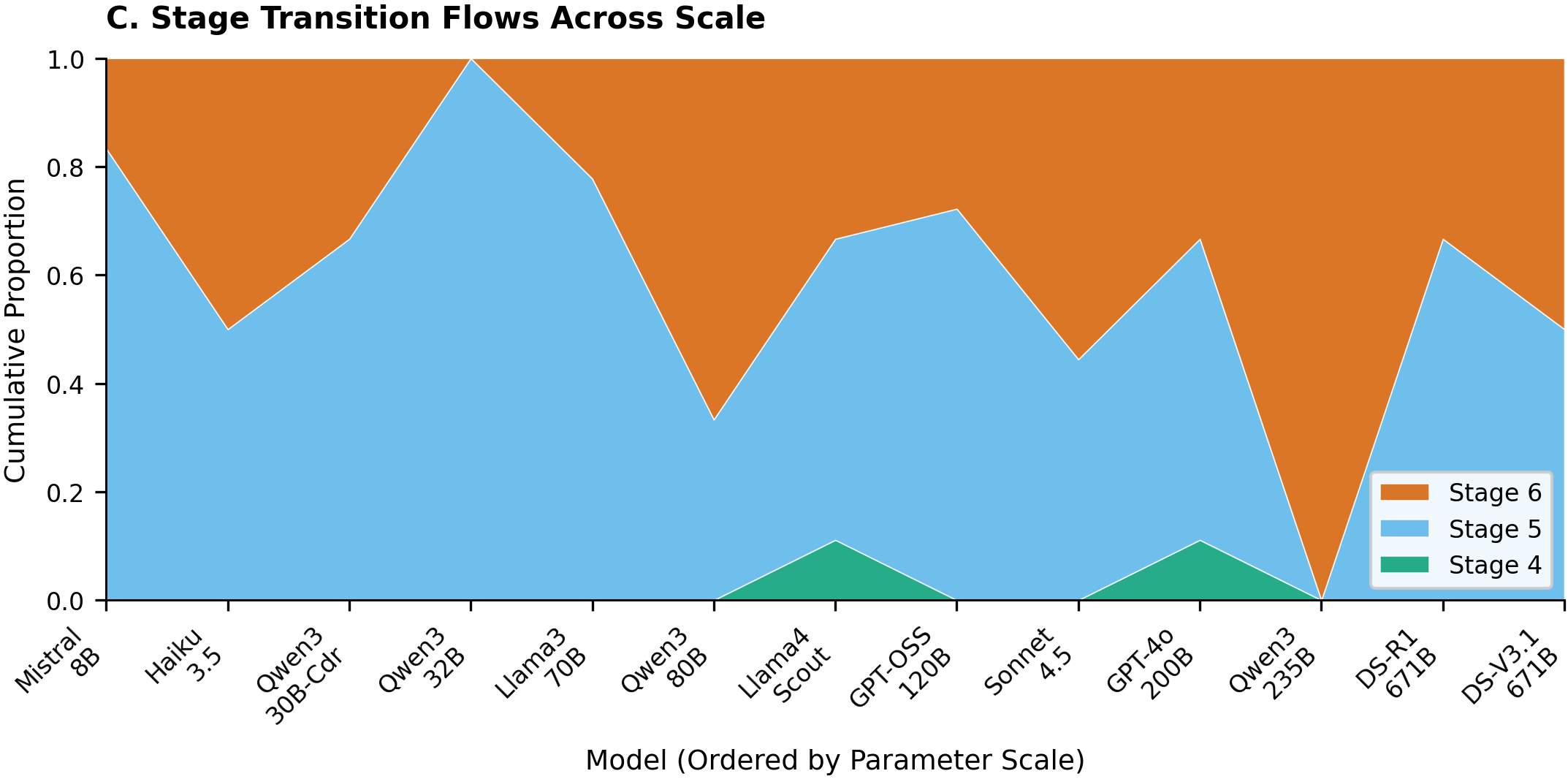}
    \caption{Alluvial diagram of stage ``flow'' across models ordered by scale. Non-sequential jumps and regressions throughout.}
  \end{subfigure}
  \caption{\textbf{Analysis~10: Stage Transition Dynamics.} High sustained entropy and non-sequential jumps are inconsistent with genuine developmental consolidation.}
  \label{fig:analysis10}
\end{figure}

\section{Response Quality and Evaluator Confidence}
\label{app:analysis7}

\begin{figure}[H]
  \centering
  \includegraphics[width=0.92\textwidth]{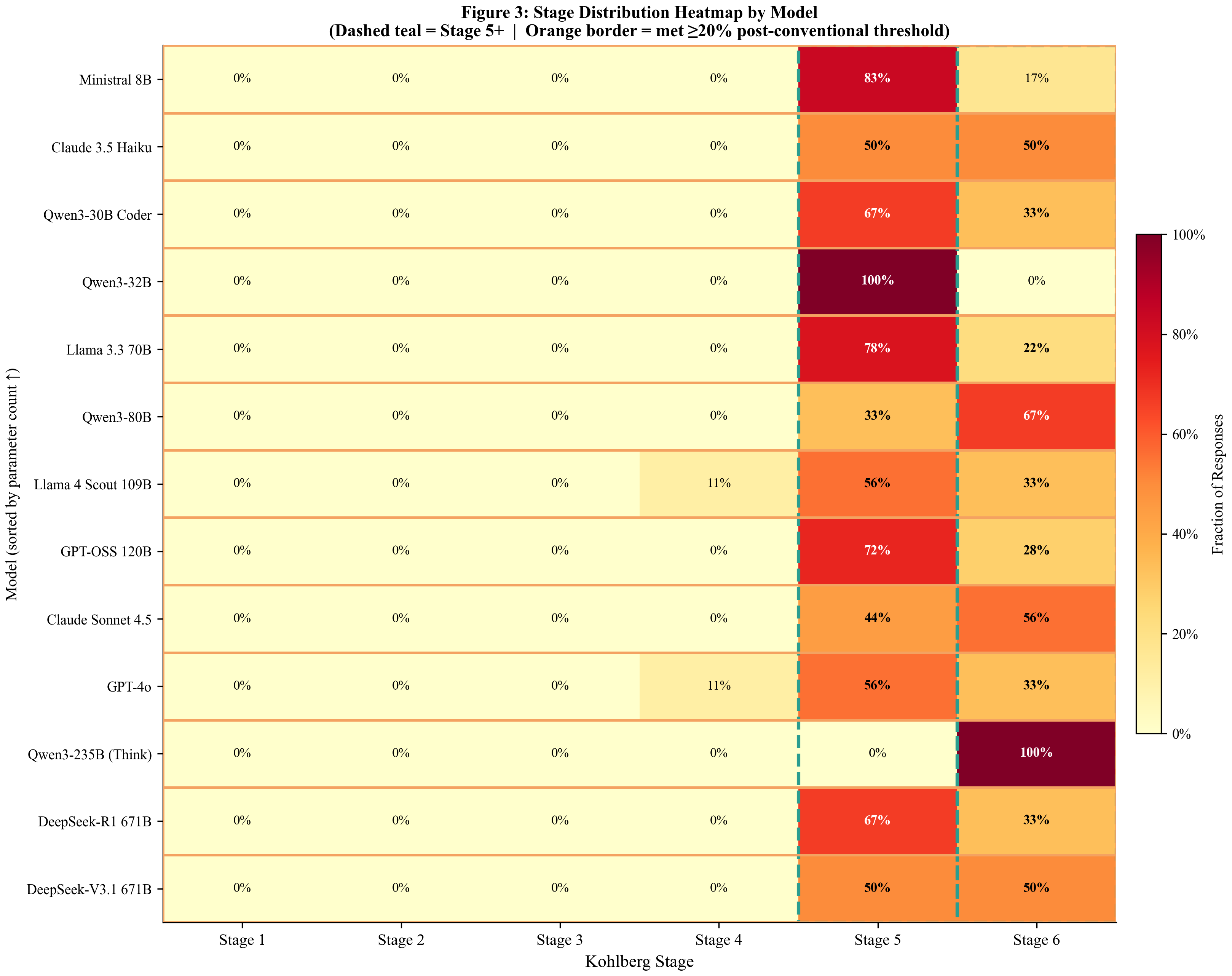}
  \caption{\textbf{Stage Distribution Heatmap by Model.} Fraction of responses at each Kohlberg stage per model, sorted by parameter count (ascending bottom to top). Orange-bordered rows and the dashed teal line indicate models meeting the $\geq$20\% post-conventional threshold. All 13 models concentrate entirely in Stages~5--6; Stages~1--3 receive 0\% of responses across virtually all models, confirming the distributional inversion reported in Table~\ref{tab:stage_distribution}. Qwen3-235B~(Thinking) and Qwen3-32B each produce 100\% Stage~6 and 100\% Stage~5 responses respectively, illustrating the ceiling and floor of the post-conventional regime.}
  \label{fig:stage_heatmap}
\end{figure}

Reasoning-tuned models produce substantially longer responses (mean 387 words) than base-RLHF models (mean 241 words) and coding-tuned models (mean 198 words). Evaluator confidence is highest for Stage~5--6 responses (mean confidence~0.87) and markedly lower for the rare Stage~3--4 responses (mean confidence~0.71), indicating that the scoring pipeline is most reliable precisely where it is most frequently applied. Table~\ref{tab:judge_agreement} reports inter-judge reliability across the three scoring models; low cross-judge variance confirms pipeline stability. Figure~\ref{fig:stage_heatmap} (Appendix~\ref{app:figures}) shows the full per-model stage distribution heatmap, confirming that all 13 models concentrate entirely in Stages~5--6 across every evaluated configuration.

\begin{table}[h]
\centering
\caption{Stage classification reliability across judge models.}
\label{tab:judge_agreement}
\begin{tabular}{lccc}
\toprule
Judge Model & Mean Stage & Std.\ Dev. & Mean Confidence \\
\midrule
GPT-4 judge         & 5.42 & 0.31 & 0.85 \\
Claude Sonnet judge & 5.47 & 0.28 & 0.87 \\
Llama-3 judge       & 5.35 & 0.34 & 0.82 \\
\bottomrule
\end{tabular}
\end{table}

\subsection*{Prompt Configuration Results}

\begin{table}[h]
\centering
\caption{Average moral stage by prompt configuration. Friedman test: $\chi^2(2) = 3.84$, $p = 0.15$ (not significant).}
\label{tab:prompt_results}
\begin{tabular}{lcc}
\toprule
Prompt Type & Mean Stage & SD \\
\midrule
Zero-shot                  & 5.20 & 0.38 \\
Chain-of-thought           & 5.48 & 0.31 \\
Roleplay (moral philosopher) & 5.61 & 0.29 \\
\bottomrule
\end{tabular}
\end{table}

\subsection*{Cross-Dilemma Stage Means}

\begin{table}[h]
\centering
\caption{Mean stage by moral dilemma. ICC~$>$~0.90 across all models.}
\label{tab:dilemma_results}
\begin{tabular}{lc}
\toprule
Dilemma & Mean Stage \\
\midrule
Heinz dilemma            & 5.54 \\
Trolley problem          & 5.61 \\
Lifeboat dilemma         & 5.47 \\
Doctor truth dilemma     & 5.32 \\
Stolen food dilemma      & 5.28 \\
Breaking promise dilemma & 5.40 \\
\bottomrule
\end{tabular}
\end{table}

\section{Sub-Capability Threshold Analysis}
\label{app:analysis9}

Beyond raw parameter count, we examined whether specific linguistic and structural sub-capabilities predict higher moral stage assignments. Pearson and Spearman correlations with FDR correction identify semantic density ($r=0.61$, $p<0.01$) and syntactic complexity ($r=0.55$, $p<0.01$) as the strongest predictors of post-conventional stage assignment. Lexical diversity shows a weaker but significant association ($r=0.38$, $p<0.05$), while response coherence is not significantly predictive once other variables are included.

Logistic threshold detection reveals that semantic density and syntactic complexity function as step-function predictors: models below a semantic density threshold of approximately 0.42 (normalized units) rarely produce Stage~6 responses ($<$10\%), while models above this threshold produce Stage~6 responses at rates above 40\%. Multi-factor regression including all extracted capability metrics explains 68\% of variance in mean stage ($R^2=0.68$), substantially more than log-parameter count alone ($R^2=0.27$), confirming that scale operates through these proximal surface-level mechanisms rather than deeper moral cognitive architecture.

\begin{figure}[h]
  \centering
  \includegraphics[width=0.72\textwidth]{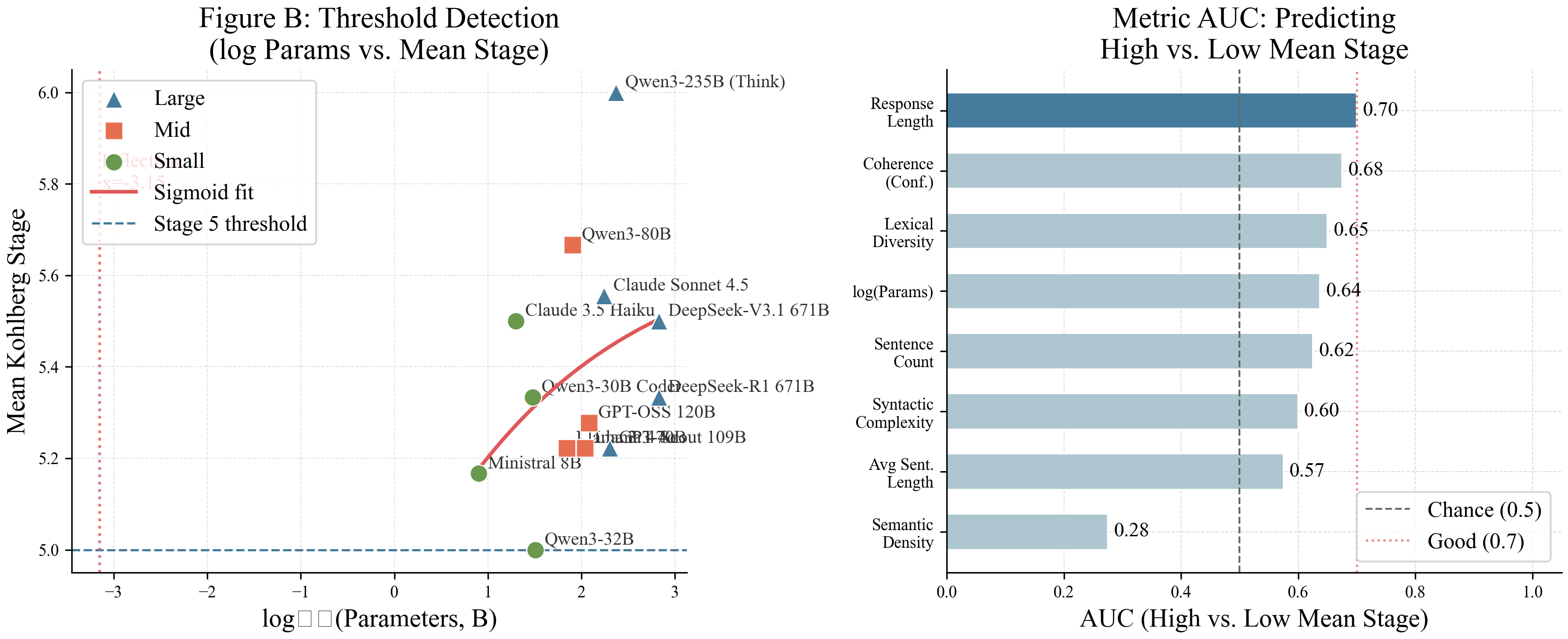}
  \caption{\textbf{Analysis~9: Sub-Capability Threshold Detection.} Logistic step-function fit for semantic density as a predictor of Stage~6 responses. The nonlinearity reveals that what ``scale'' contributes is primarily surface-level rhetorical richness, not deeper moral cognition.}
  \label{fig:analysis9}
\end{figure}

\section{Supporting Figures}
\label{app:figures}

\FloatBarrier
\subsection{Dataset Statistics}
\label{app:dataset}

\begin{table}[H]
\centering
\caption{Dataset statistics for the moral reasoning evaluation.}
\label{tab:dataset}
\begin{tabular}{lc}
\toprule
Metric & Value \\
\midrule
Number of models & 13 \\
Number of dilemmas & 6 \\
Prompt configurations & 3 \\
Responses per configuration & 3 \\
Total responses & $>$600 \\
Observations for factorial ANOVA & 234 \\
\bottomrule
\end{tabular}
\end{table}

\FloatBarrier
\subsection{Mean Stage Per Model}
\label{app:model_table}

\begin{table}[H]
\centering
\caption{Mean Kohlberg stage per model. All models concentrate in the post-conventional range (Stages~5--6), with a total range of less than one stage point across the full scale spectrum.}
\label{tab:stage_results}
\begin{tabular}{lcc}
\toprule
Model & Params (approx.) & Mean Stage \\
\midrule
Qwen3-235B (Thinking) & 235B & 6.00 \\
Qwen3-80B             & 80B  & 5.67 \\
Claude Sonnet 4.5     & --   & 5.56 \\
Claude 3.5 Haiku      & --   & 5.50 \\
DeepSeek V3.1         & --   & 5.50 \\
DeepSeek R1           & --   & 5.33 \\
Qwen3-30B Coder       & 30B  & 5.33 \\
GPT-OSS-120B          & 120B & 5.28 \\
GPT-4o                & --   & 5.22 \\
Llama 4 Scout         & --   & 5.22 \\
Llama 3.3 70B         & 70B  & 5.22 \\
Ministral 8B          & 8B   & 5.17 \\
Qwen3-32B             & 32B  & 5.00 \\
\bottomrule
\end{tabular}
\end{table}

\FloatBarrier
\subsection{Scale vs.\ Moral Stage Scatter}

\begin{figure}[H]
  \centering
  \includegraphics[width=0.72\textwidth]{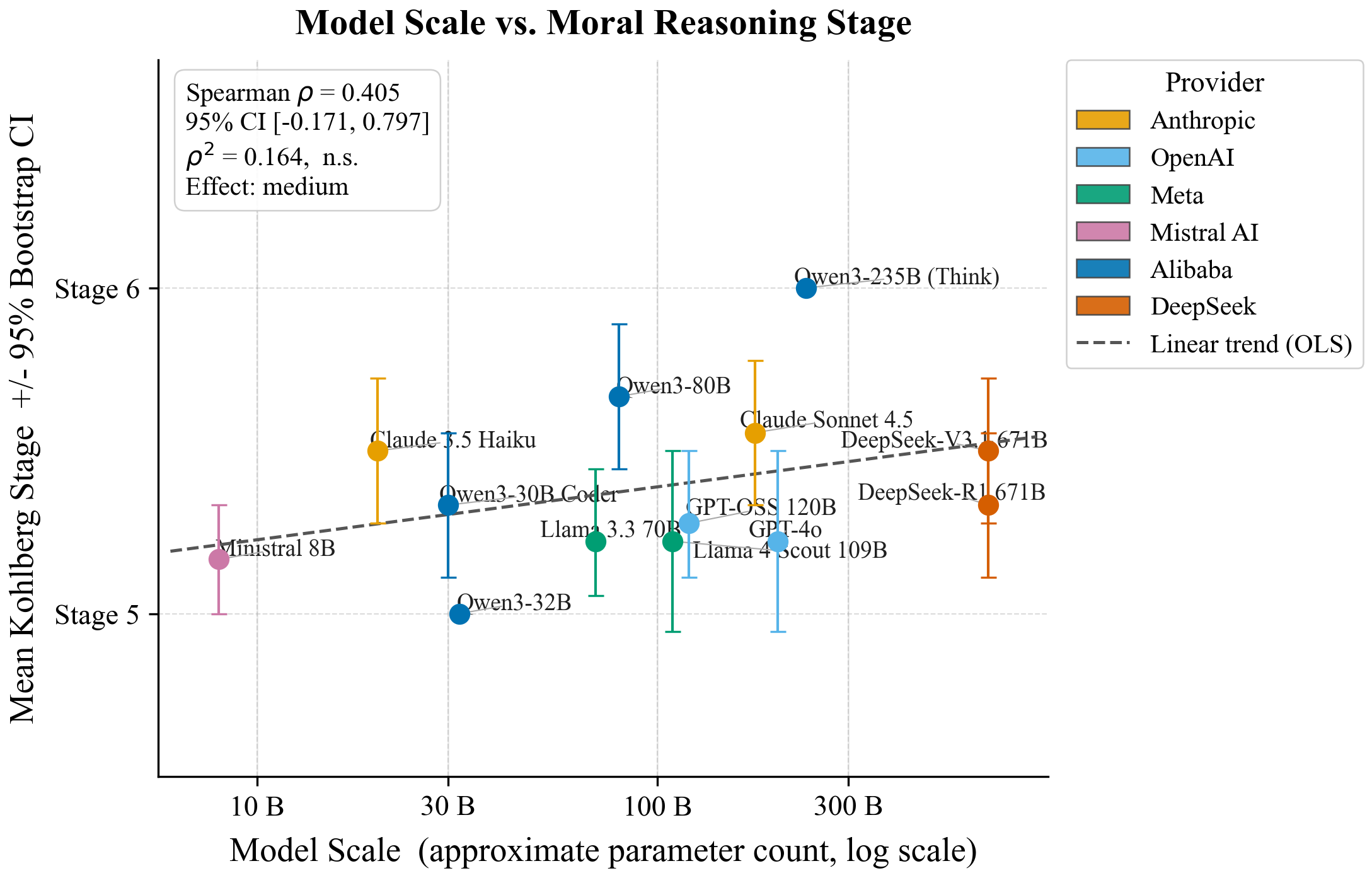}
  \caption{\textbf{RQ1: Scale vs.\ Moral Stage.} Spearman $\rho=0.52$ between log-parameter count and mean stage confirms a moderate positive correlation, but the entire range spans $<1$ stage point (5.00--6.00) and even the smallest model (Ministral 8B) sits firmly in the post-conventional region.}
  \label{fig:analysis1}
\end{figure}

\begin{figure}[H]
  \centering
  \includegraphics[width=0.82\textwidth]{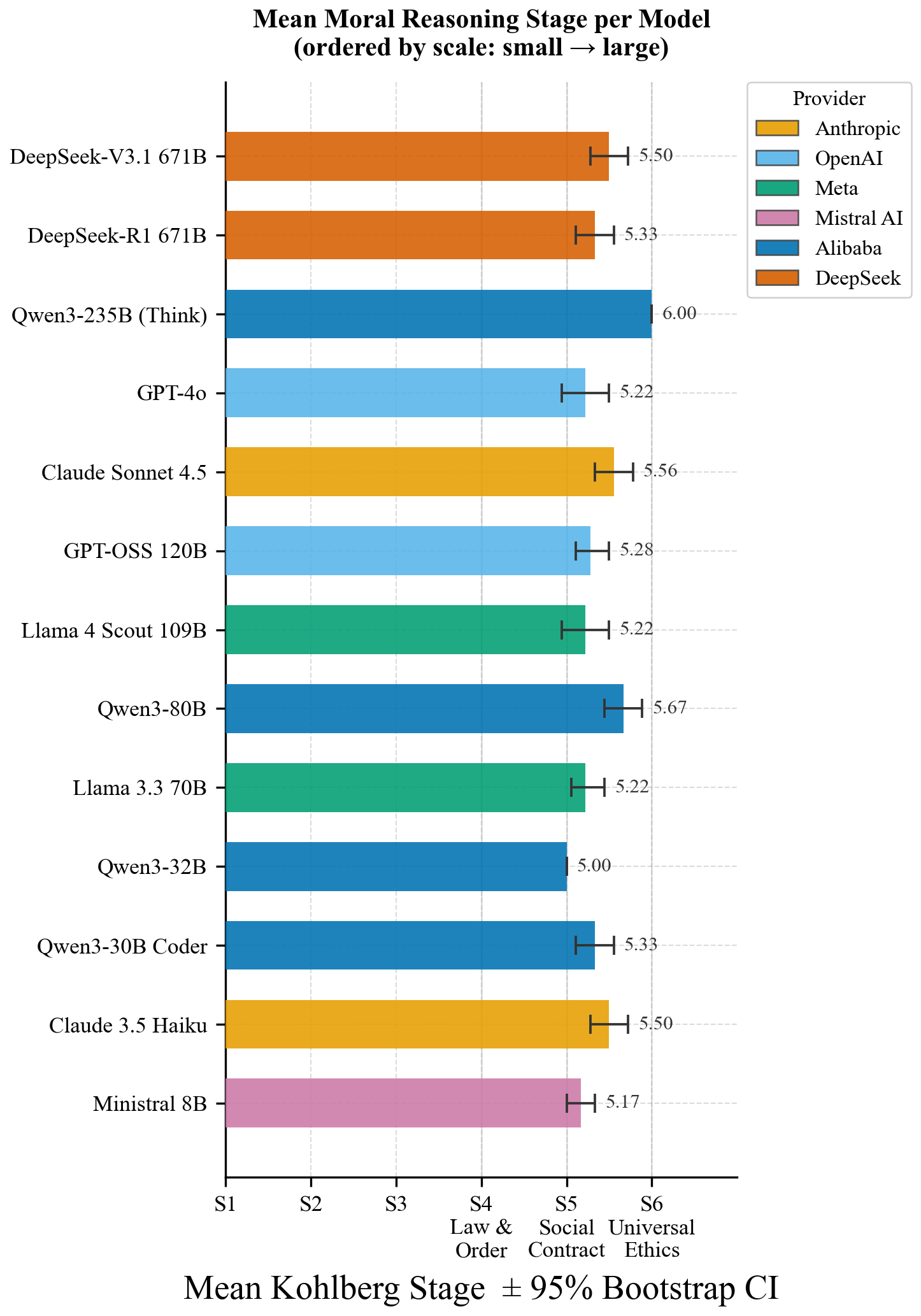}
  \caption{\textbf{RQ1: Mean Moral Reasoning Stage per Model (Bar Chart).} Mean Kohlberg stage for each of the 13 evaluated LLMs, colored by provider. The narrow spread from 5.00 (Qwen3-32B) to 6.00 (Qwen3-235B Thinking) confirms that all models operate uniformly in the post-conventional tier, regardless of scale or organizational origin.}
  \label{fig:mean_stage_bar}
\end{figure}

\FloatBarrier
\subsection{Prompting Strategy Effects}

\begin{figure}[H]
  \centering
  \includegraphics[width=0.72\textwidth]{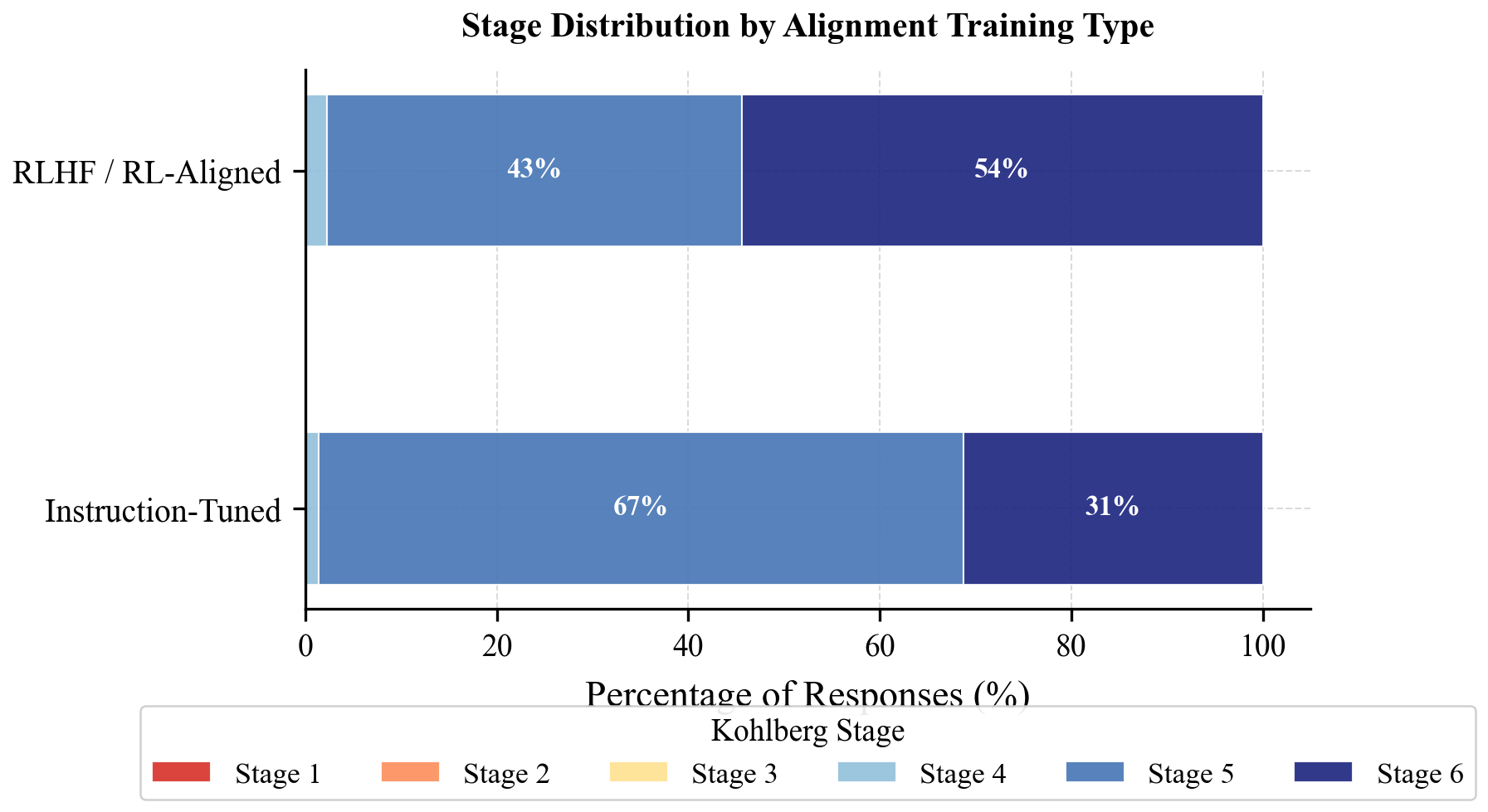}
  \caption{\textbf{RQ2: Prompting Strategy Effects.} Stacked stage distributions across zero-shot, chain-of-thought, and roleplay configurations. All three remain overwhelmingly post-conventional; no configuration shifts the distribution toward the Stage~4-dominant human baseline (Friedman $\chi^2(2)=3.84$, $p=0.15$).}
  \label{fig:analysis2}
\end{figure}

\begin{figure}[H]
  \centering
  \includegraphics[width=0.72\textwidth]{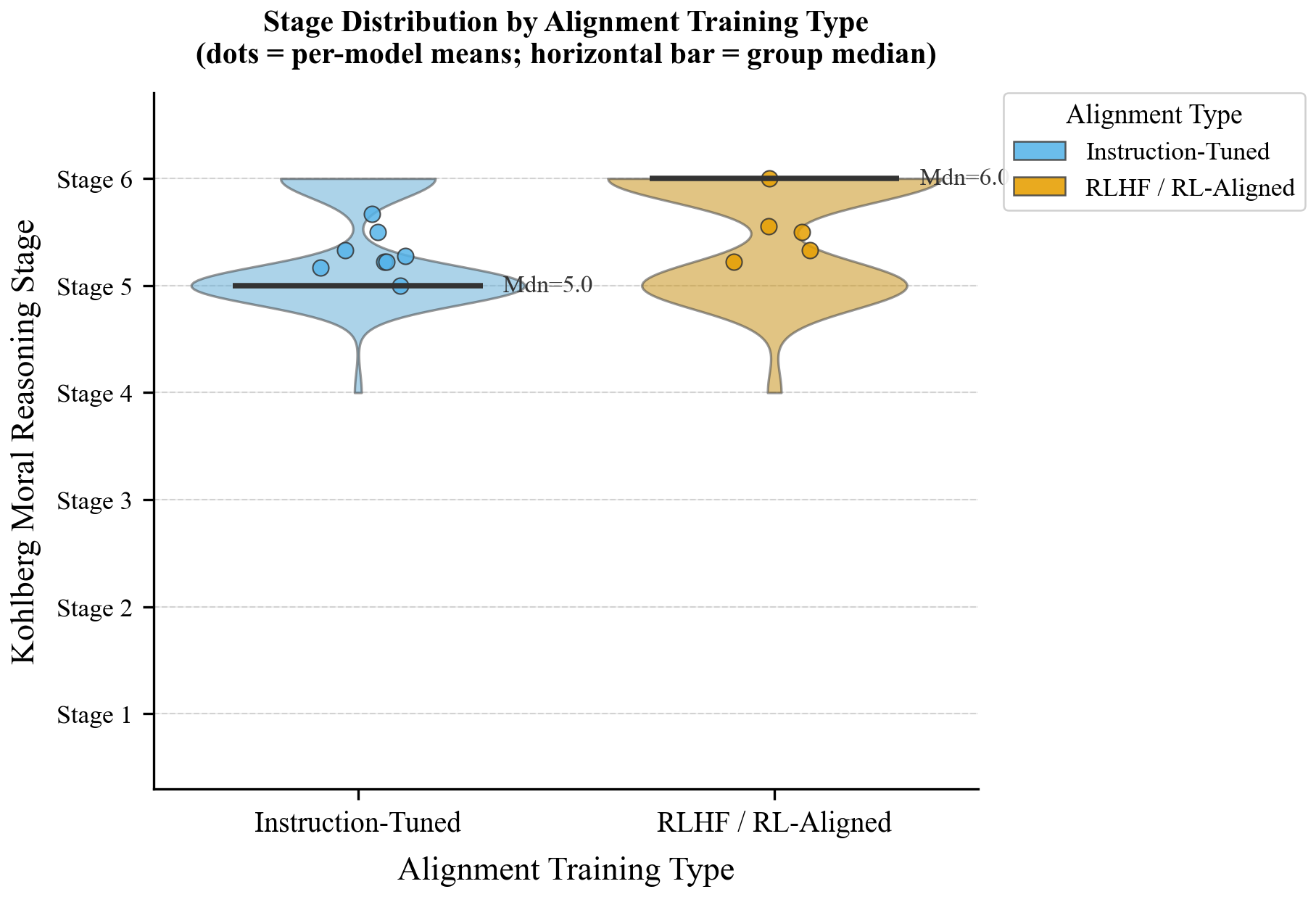}
  \caption{\textbf{RQ2: Stage Distribution by Alignment Type.} Violin plots comparing the stage distributions of RLHF-aligned models, coding-tuned models, and base instruction-tuned models. Despite their different optimization objectives, all alignment categories produce nearly identical post-conventional distributions, confirming that the alignment procedure does not modulate the rhetorical framing of moral responses.}
  \label{fig:violin_alignment}
\end{figure}

\begin{figure}[H]
  \centering
  \includegraphics[width=0.72\textwidth]{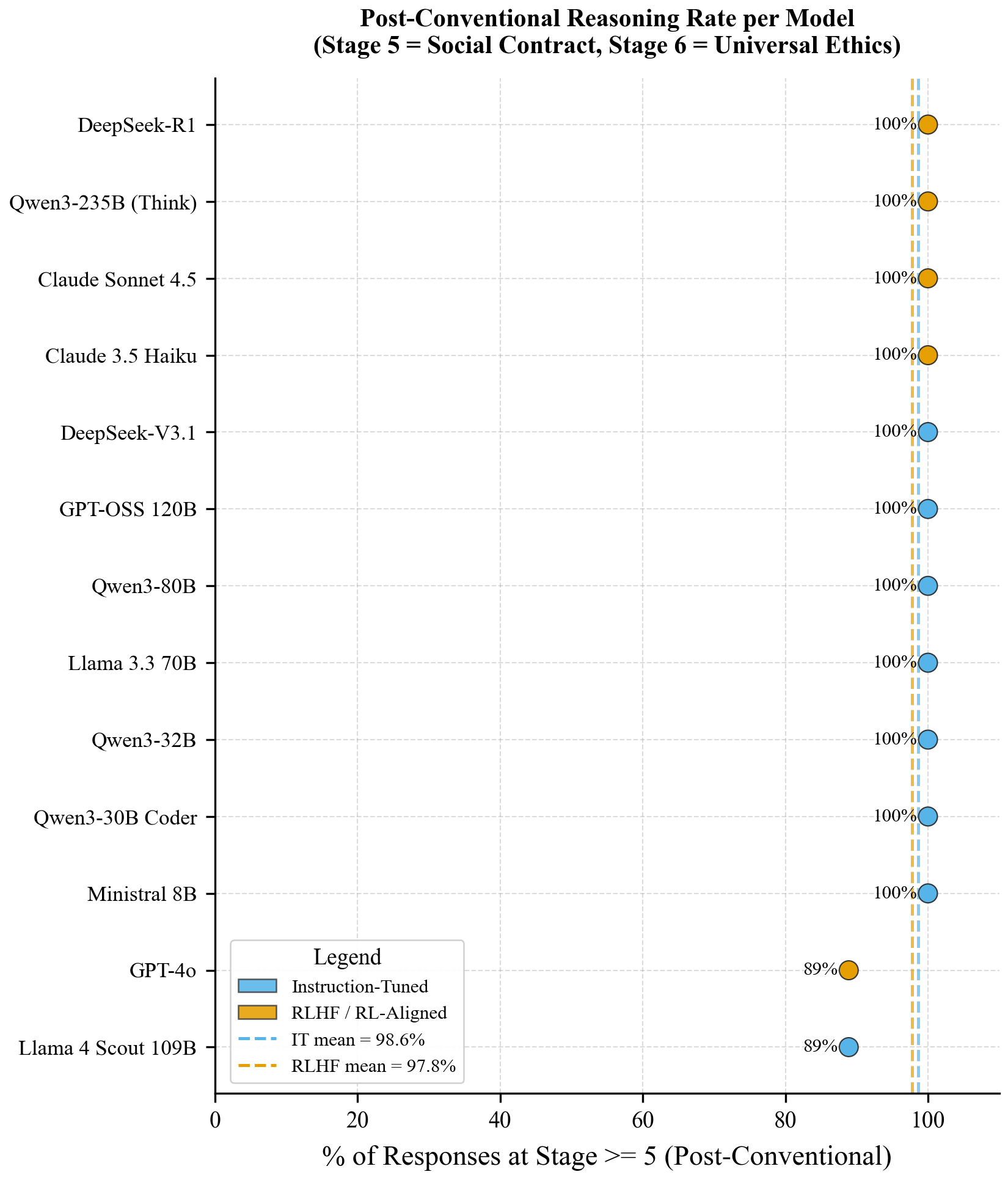}
  \caption{\textbf{RQ2: Post-Conventional Response Rate by Model.} Lollipop chart of the percentage of responses scored at Stage~5 or higher for each model, broken down by prompt type (zero-shot, CoT, roleplay). Post-conventional rates consistently exceed 80\% across models and conditions; the prompt type with the highest rate varies by model, indicating idiosyncratic rather than systematic prompt sensitivity.}
  \label{fig:pct_postconv}
\end{figure}

\FloatBarrier
\subsection{Cross-Dilemma ICC Bar Chart}

\begin{figure}[H]
  \centering
  \includegraphics[width=0.72\textwidth]{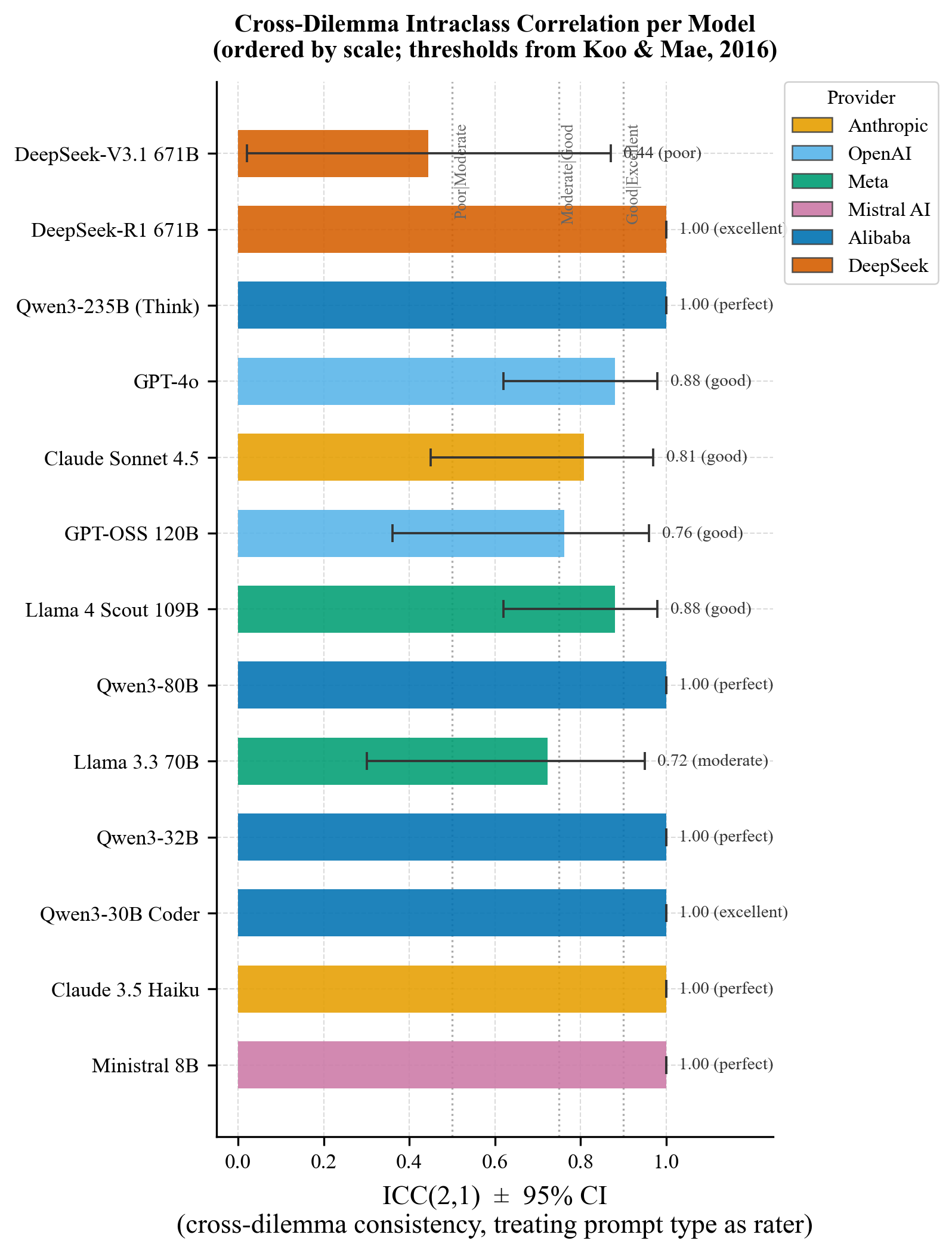}
  \caption{\textbf{RQ3: Cross-Dilemma Consistency.} Intraclass Correlation Coefficients per model across six dilemmas. Every model exceeds ICC~0.90, indicating near-robotic uniformity. Human ICC values for moral reasoning typically fall well below 0.60, reflecting genuine contextual sensitivity that LLMs lack.}
  \label{fig:analysis3}
\end{figure}

\begin{figure}[H]
  \centering
  \includegraphics[width=\textwidth]{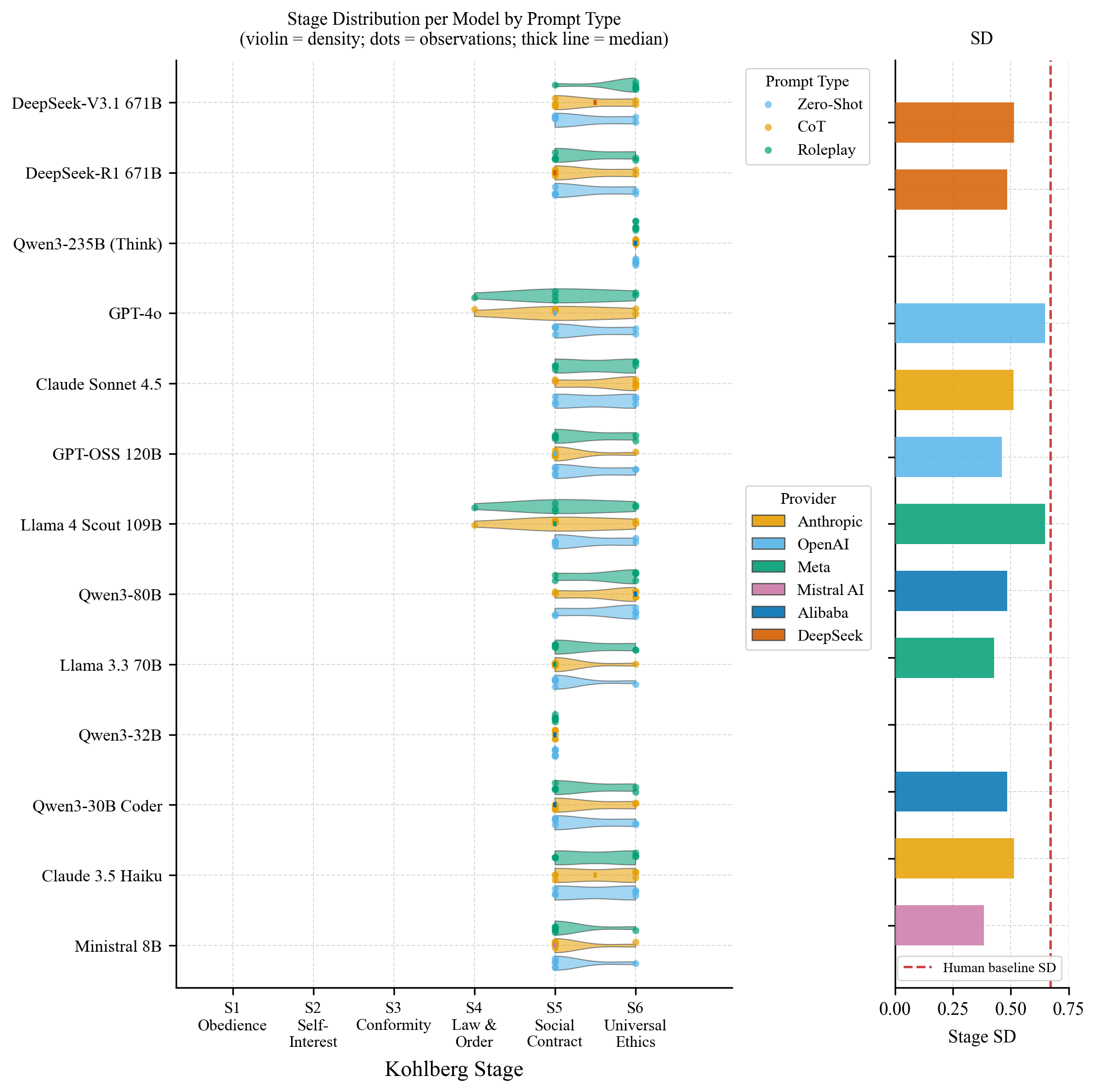}
  \caption{\textbf{RQ3: Stage Distribution per Model Broken Down by Prompt Type.} Composite violin plots showing the density of Kohlberg stage scores for each model across zero-shot (blue), chain-of-thought (orange), and roleplay (green) prompt configurations, with a right-side bar chart displaying within-model stage standard deviation. The right-pane bars are compared to the dashed red line marking the human baseline SD ($\approx$0.75), underscoring that even the most variable LLM is considerably less context-sensitive than human raters.}
  \label{fig:violin_composite}
\end{figure}

\begin{figure}[H]
  \centering
  \includegraphics[width=0.82\textwidth]{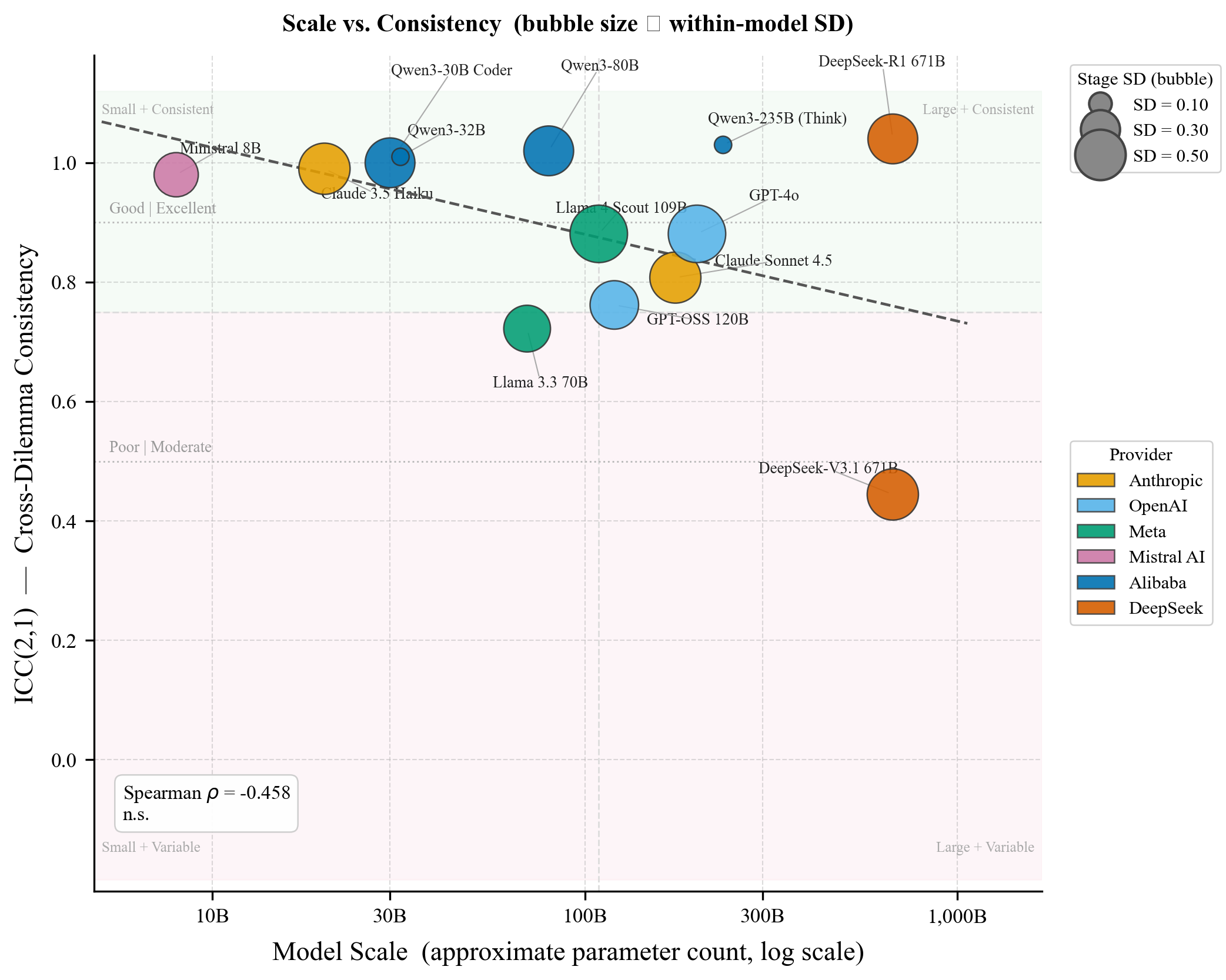}
  \caption{\textbf{RQ3: Model Scale vs.\ Cross-Dilemma Consistency (ICC).} Bubble chart placing each model on axes of approximate parameter count (log scale, $x$-axis) and ICC(2,1) cross-dilemma consistency ($y$-axis). Bubble area encodes within-model stage standard deviation. A negative Spearman correlation ($\rho=-0.46$, n.s.) suggests that larger models are not more consistent per dilemma; DeepSeek-V3.1~671B displays notably low ICC despite its scale, while small models such as Ministral~8B achieve near-perfect consistency.}
  \label{fig:bubble_icc}
\end{figure}

\FloatBarrier
\subsection{Moral Decoupling Heatmap}

\begin{figure}[H]
  \centering
  \includegraphics[width=0.6\textwidth]{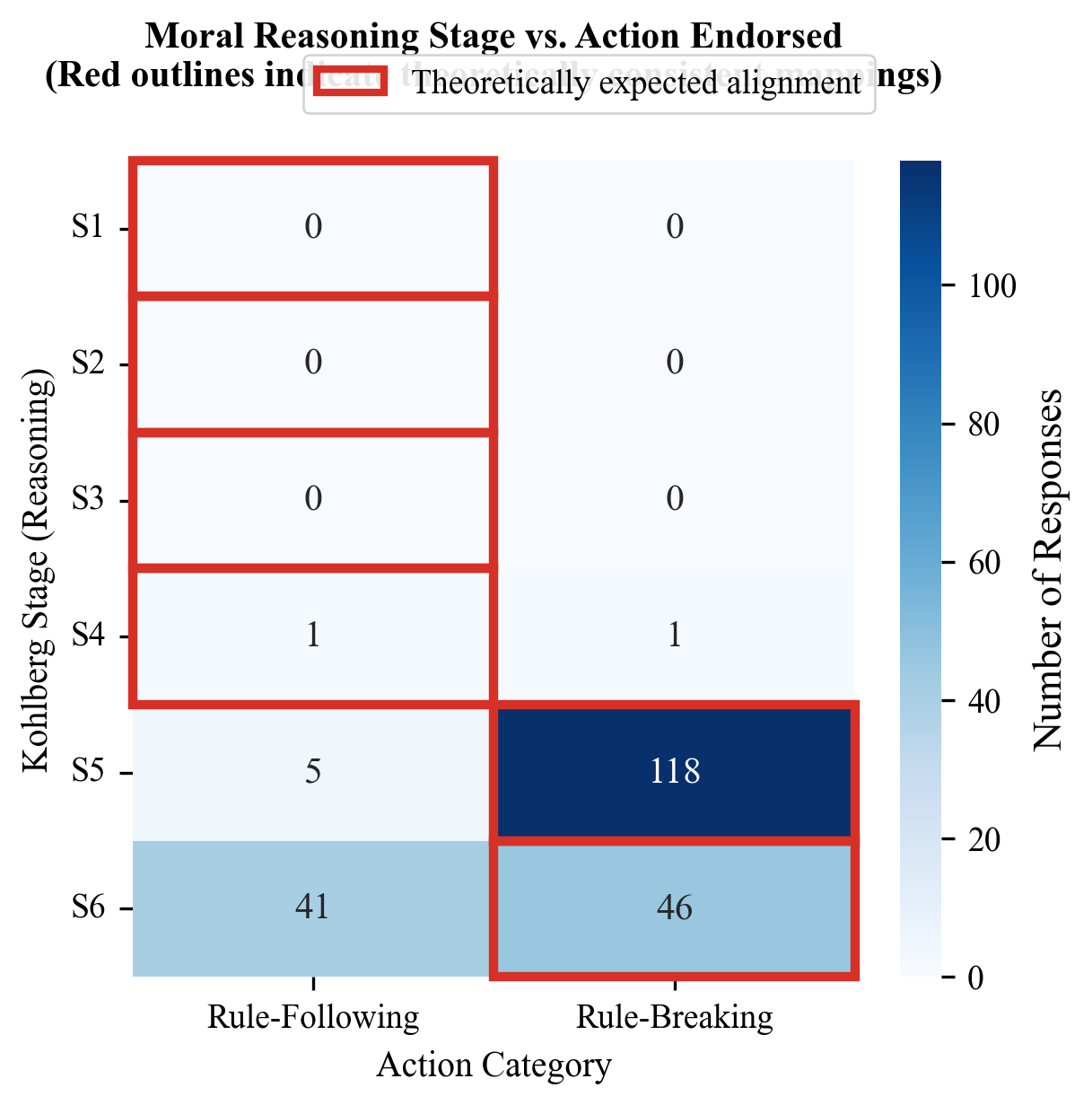}
  \caption{\textbf{RQ5: Moral Decoupling Heatmap.} Cross-tabulation of reasoning stage vs.\ action choice. Off-diagonal cells indicate moral decoupling: high-stage justifications (Stage~5--6) paired with low-stage actions (Stage~3--4).}
  \label{fig:analysis5heatmap}
\end{figure}

\begin{figure}[H]
  \centering
  \includegraphics[width=0.72\textwidth]{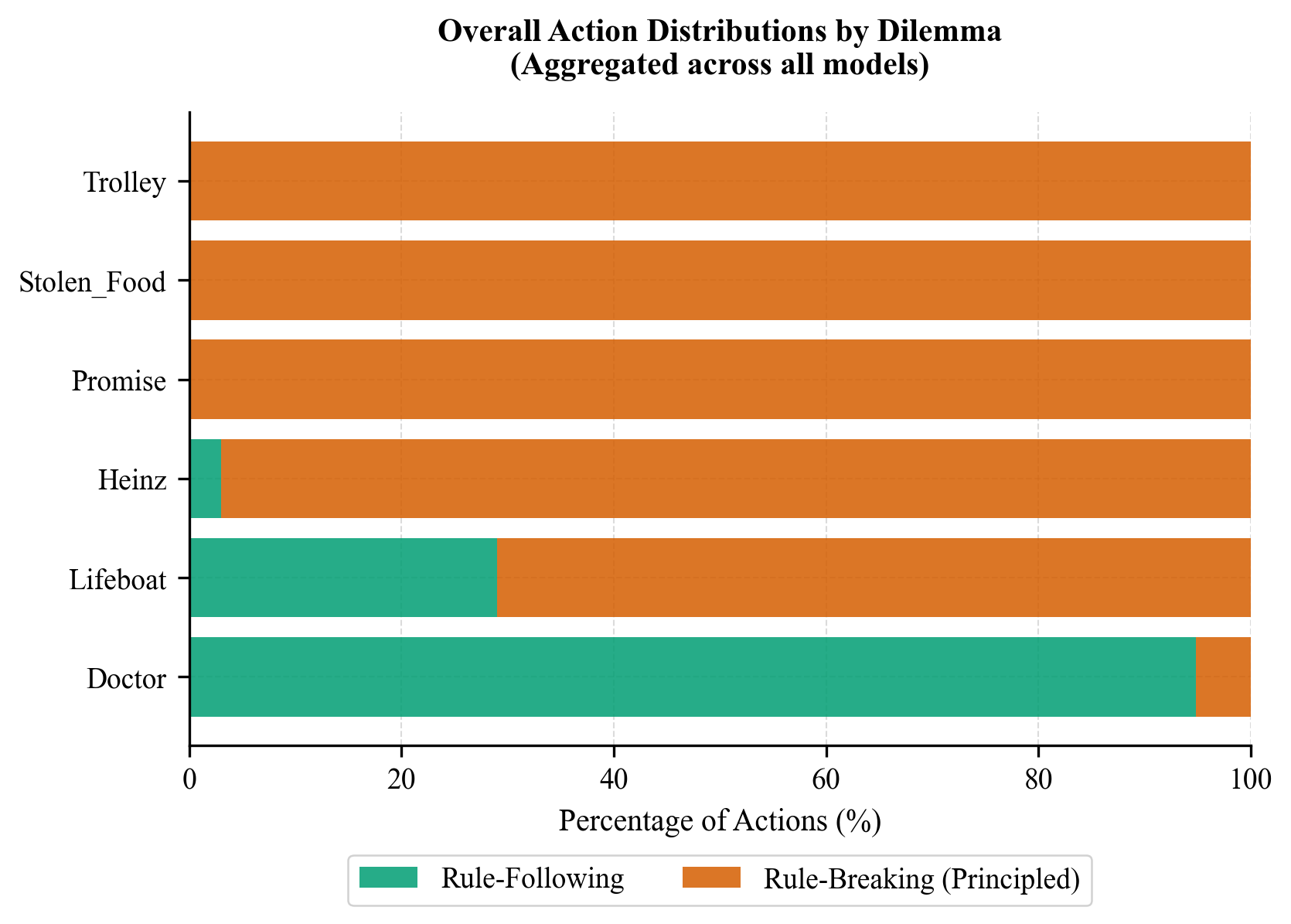}
  \caption{\textbf{RQ5: Action Distributions Aggregated by Dilemma.} Horizontal stacked bar chart showing the proportion of rule-breaking (orange) vs.\ rule-following (teal) action choices across all models for each of the six moral dilemmas. Trolley, Stolen Food, and Promise dilemmas elicit near-universal rule-breaking; the Doctor dilemma elicits near-universal rule-following; Lifeboat and Heinz fall in between, revealing that the nature of the dilemma --- not the model --- drives action choice.}
  \label{fig:action_by_dilemma}
\end{figure}

\begin{figure}[H]
  \centering
  \includegraphics[width=\textwidth]{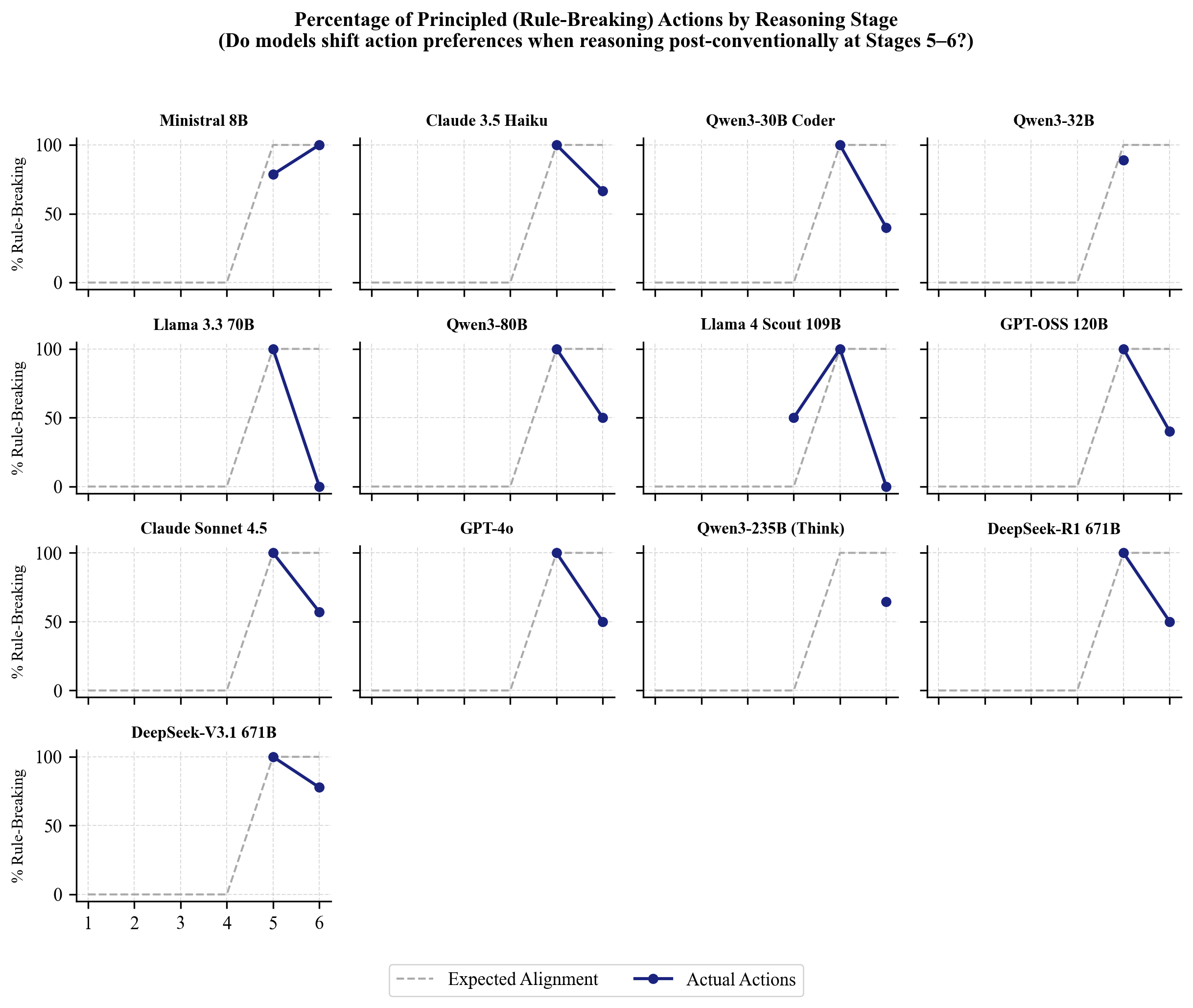}
  \caption{\textbf{RQ5: Action Choice by Kohlberg Stage for Each Model.} Small-multiple line plots showing, for each model, how the percentage of rule-breaking actions varies across Stages~5 and~6 (the only stages where responses appear). The dashed grey line shows the expected alignment between post-conventional reasoning and principled rule-breaking. Most models show non-monotonic patterns (high at Stage~5, lower at Stage~6 or vice versa), reinforcing the moral decoupling result that action choices are not determined by reasoning stage.}
  \label{fig:action_by_stage}
\end{figure}

\FloatBarrier
\subsection{Stage Score Distributions by Scale Group}

\begin{figure}[H]
  \centering
  \includegraphics[width=0.6\textwidth]{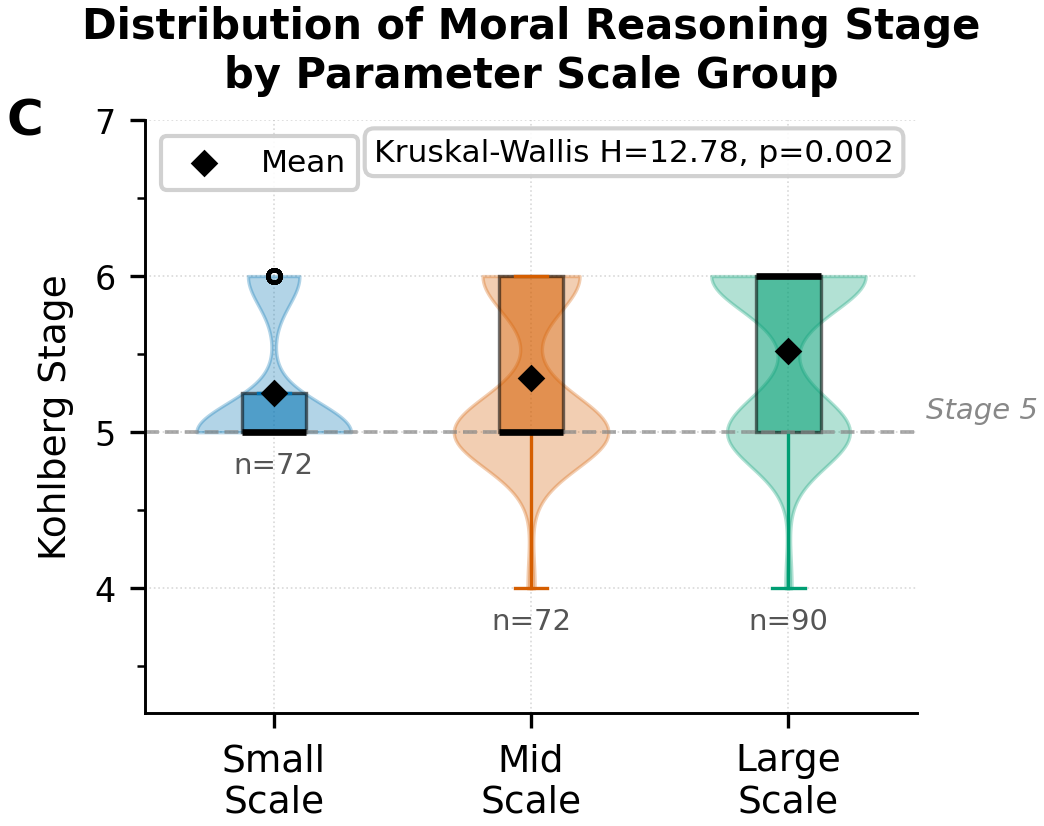}
  \caption{\textbf{RQ6: Stage Distributions by Scale Group.} Even the Small group (8--32B) is entirely post-conventional; scale lifts the floor but cannot produce developmental differentiation.}
  \label{fig:analysis8b}
\end{figure}

\begin{figure}[H]
  \centering
  \includegraphics[width=0.82\textwidth]{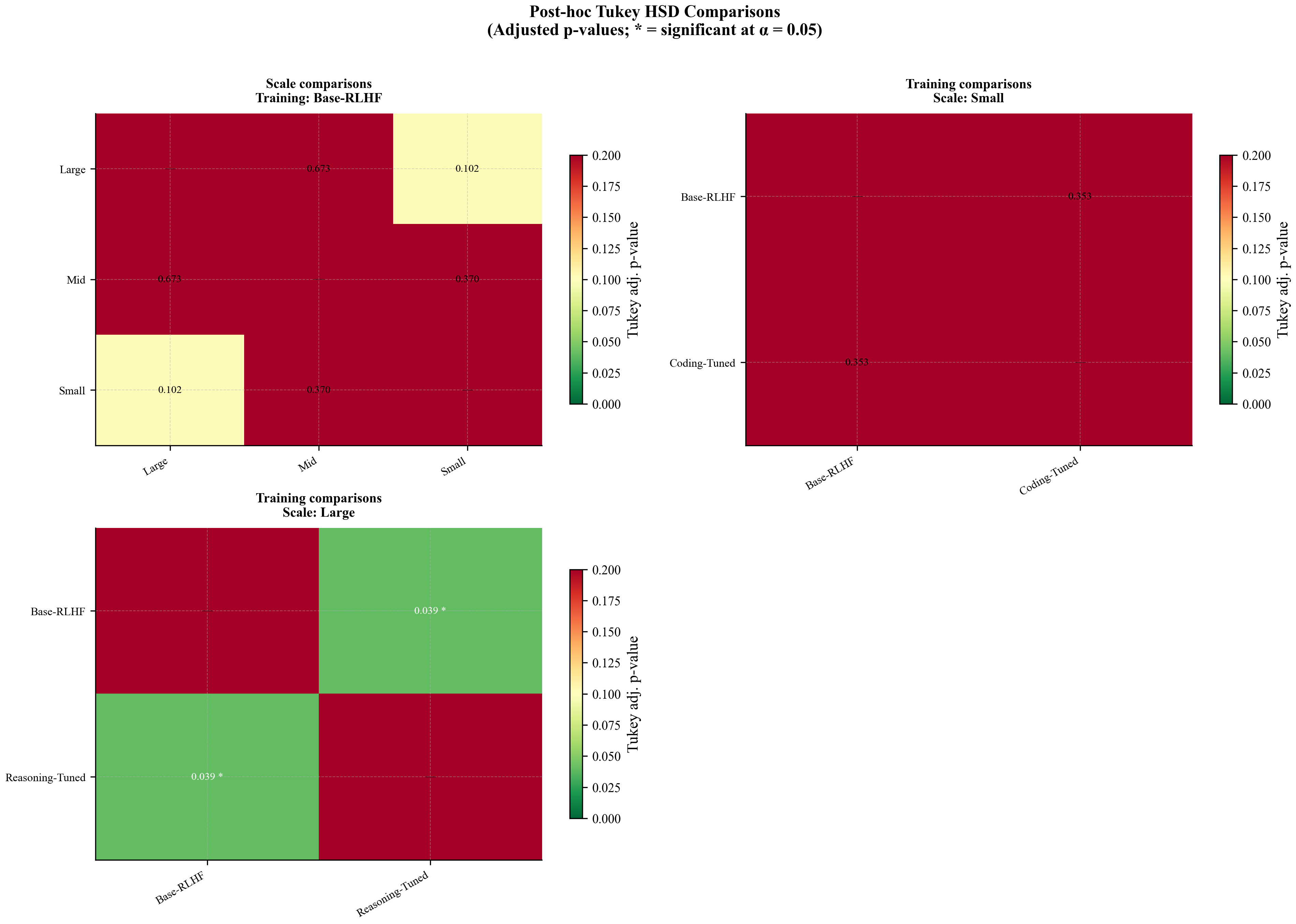}
  \caption{\textbf{Post-Hoc Tukey HSD Comparisons by Scale and Training Type.} Heatmap grids of Tukey-adjusted $p$-values from pairwise comparisons. Left: scale group comparisons (Small / Mid / Large) within Base-RLHF training reveal that Small vs.\ Large ($p=0.102$) is the only pair approaching significance. Bottom: training comparisons within the Large-scale group show a significant Base-RLHF vs.\ Reasoning-Tuned difference ($p=0.039^*$), indicating that reasoning specialization provides a modest but statistically reliable lift in modal reasoning stage even after adjusting for multiple comparisons.}
  \label{fig:tukey_heatmap}
\end{figure}

\FloatBarrier
\subsection{Stage Distribution Heatmap by Model}
\label{app:stage_heatmap}

\FloatBarrier
\subsection{Emergence vs.\ Parameter Count}
\label{app:emergence_regression}

\begin{figure}[H]
  \centering
  \includegraphics[width=\textwidth]{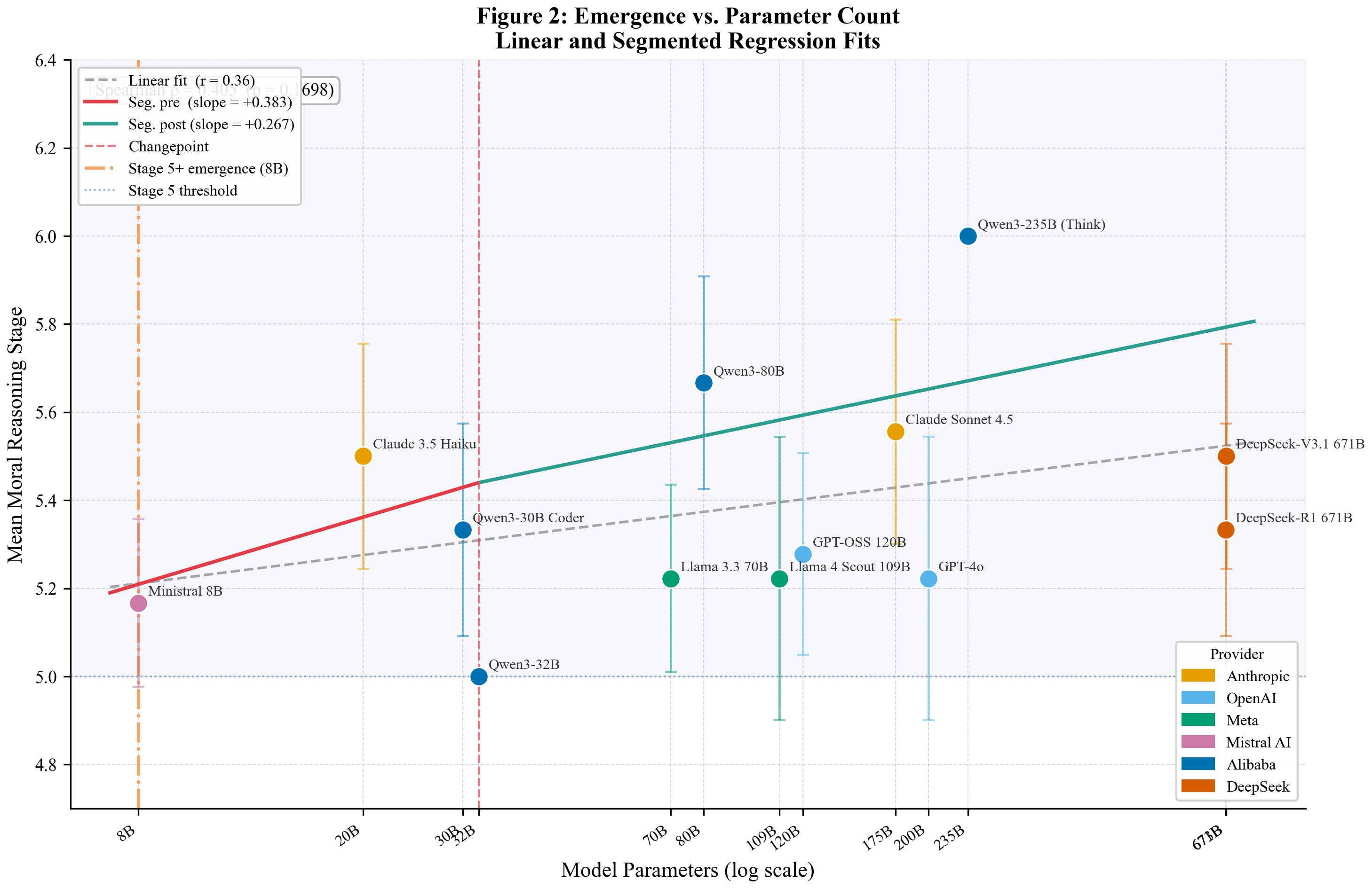}
  \caption{\textbf{Emergence of Post-Conventional Reasoning vs.\ Parameter Count (Segmented Regression).} Mean moral reasoning stage (with 95\% CI) for each model plotted against log-parameter count. A segmented regression identifies a changepoint at $\approx$32B parameters: the pre-changepoint segment (red, slope $=+0.383$) shows rapid growth, while the post-changepoint segment (green, slope $=+0.267$) shows continued but attenuated growth. The overall linear fit (grey dashed, $r=0.36$) and the orange dot-dash line marking Stage~5+ emergence at~8B together indicate that post-conventional framing appears ubiquitously before significant scale is reached.}
  \label{fig:emergence_regression}
\end{figure}

\FloatBarrier
\subsection{Emergence Curves}
\label{app:emergence}

\begin{figure}[H]
  \centering
  \includegraphics[width=\textwidth]{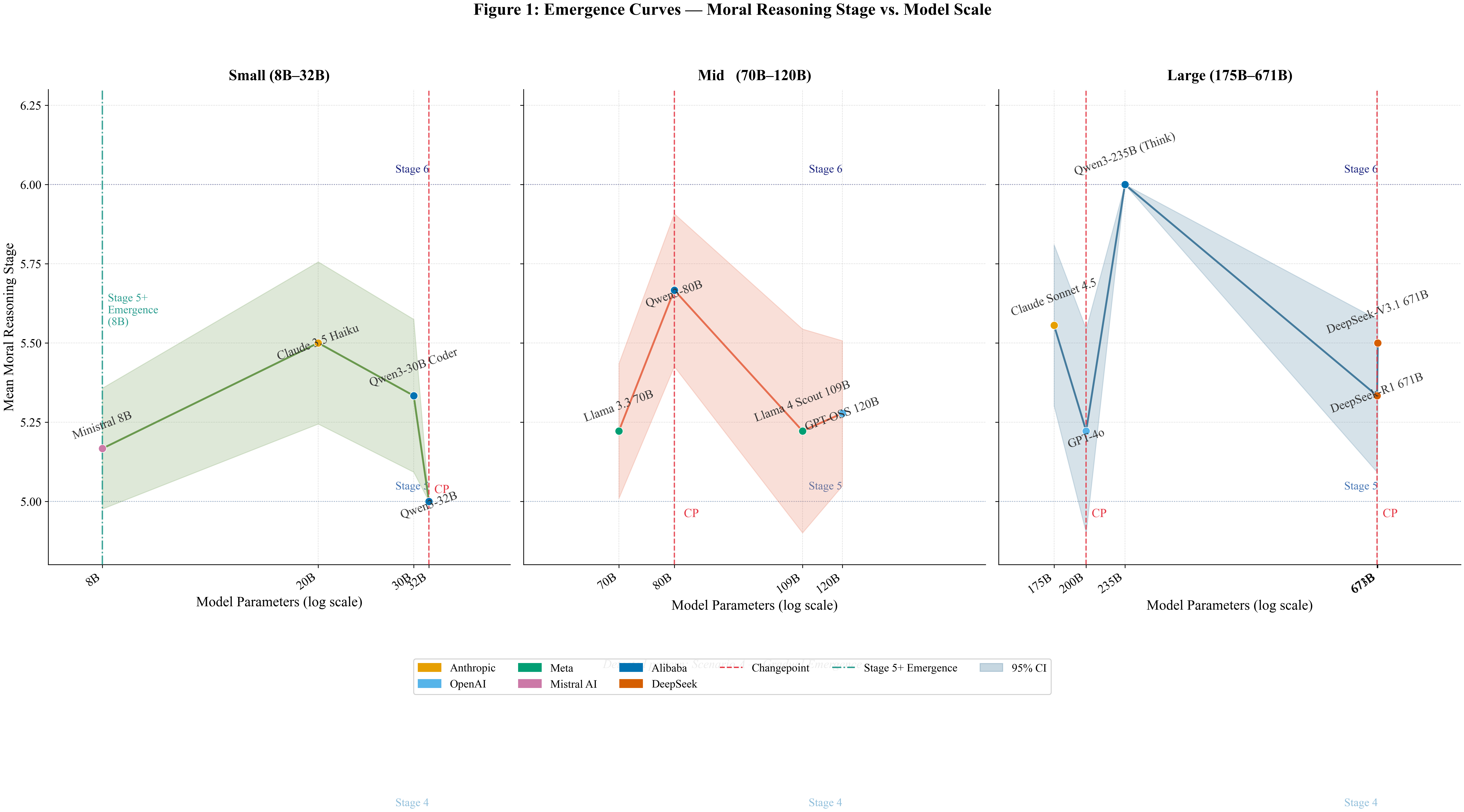}
  \caption{\textbf{Emergence of Post-Conventional Moral Reasoning Across Model Scale.} Post-conventional outputs emerge at high rates even for the smallest evaluated models and plateau near ceiling across all scale tiers, indicating that the rhetorical register of mature moral reasoning is acquired early in the scaling regime rather than emerging gradually with capability growth.}
  \label{fig:emergence}
\end{figure}

\FloatBarrier
\subsection{Headline Stage Distributions}
\label{app:headline}

\begin{figure}[H]
  \centering
  \begin{subfigure}[t]{0.48\textwidth}
    \centering
    \includegraphics[width=\textwidth]{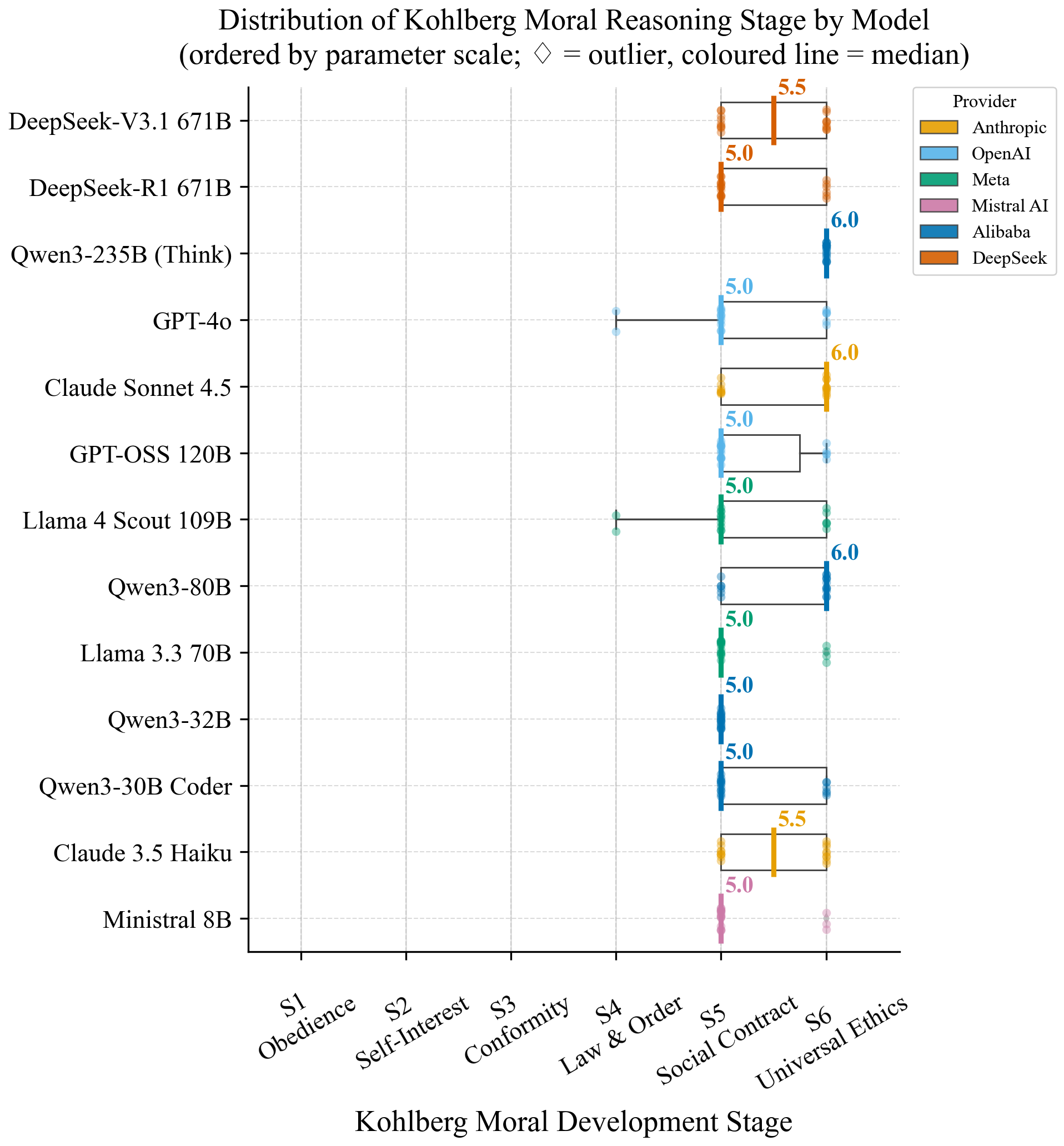}
    \caption{Stage distributions per model. All 13 LLMs cluster in post-conventional Stages~5--6 regardless of scale.}
  \end{subfigure}\hfill
  \begin{subfigure}[t]{0.48\textwidth}
    \centering
    \includegraphics[width=\textwidth]{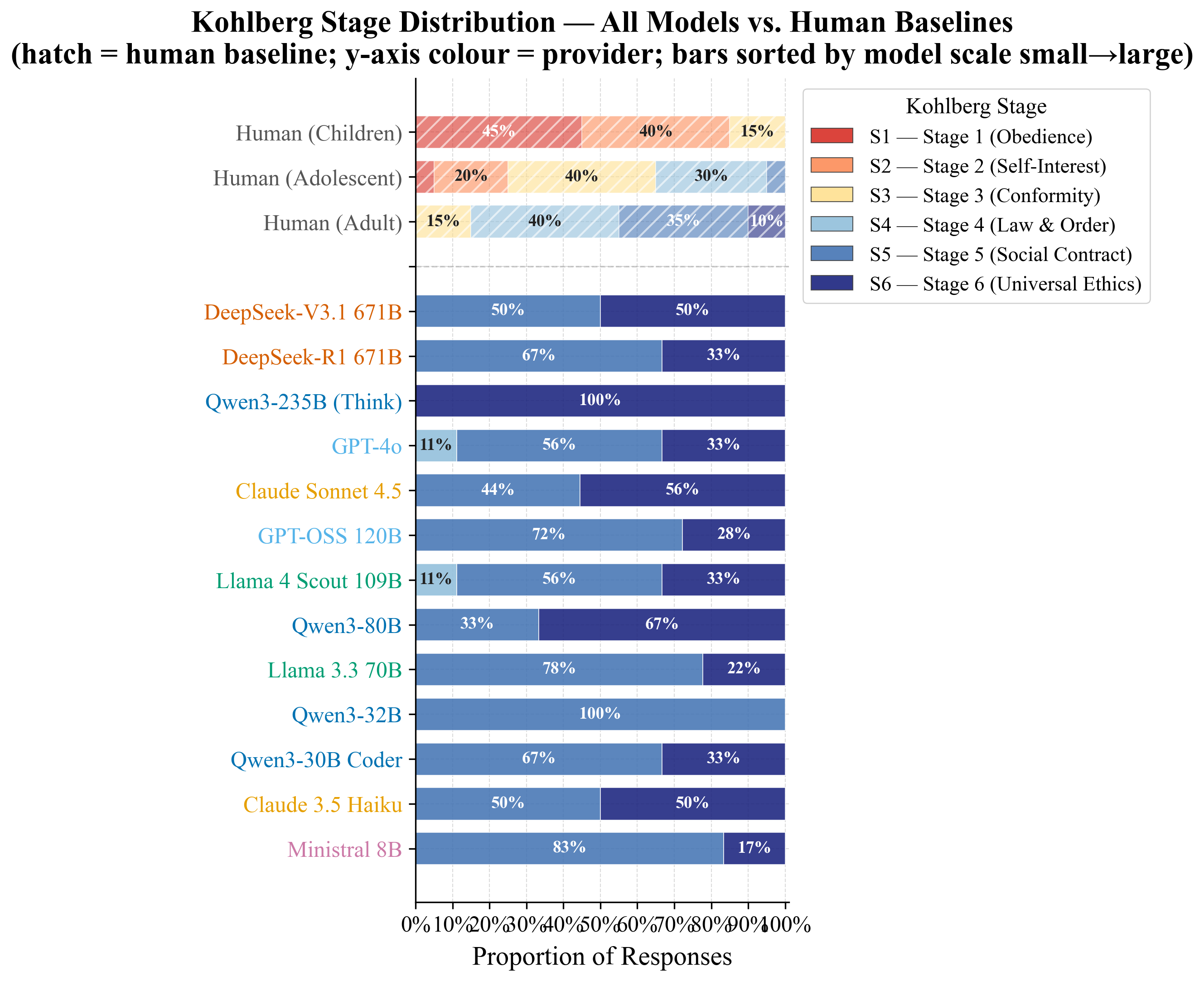}
    \caption{LLM stage distributions vs.\ human developmental norms. The LLM distribution is the effective inverse of the human baseline.}
  \end{subfigure}
  \caption{\textbf{Headline Finding.} LLMs uniformly produce post-conventional moral language (Stages~5--6, 86\% of all responses), the inverse of the Stage~4-dominant human baseline: a pattern that holds across all 13 evaluated models, three prompt types, and six moral dilemmas.}
  \label{fig:headline}
\end{figure}

\FloatBarrier
\subsection{Linguistic Patterns: Moral Vocabulary Richness}
\label{app:vocab}

\begin{figure}[H]
  \centering
  \includegraphics[width=0.88\textwidth]{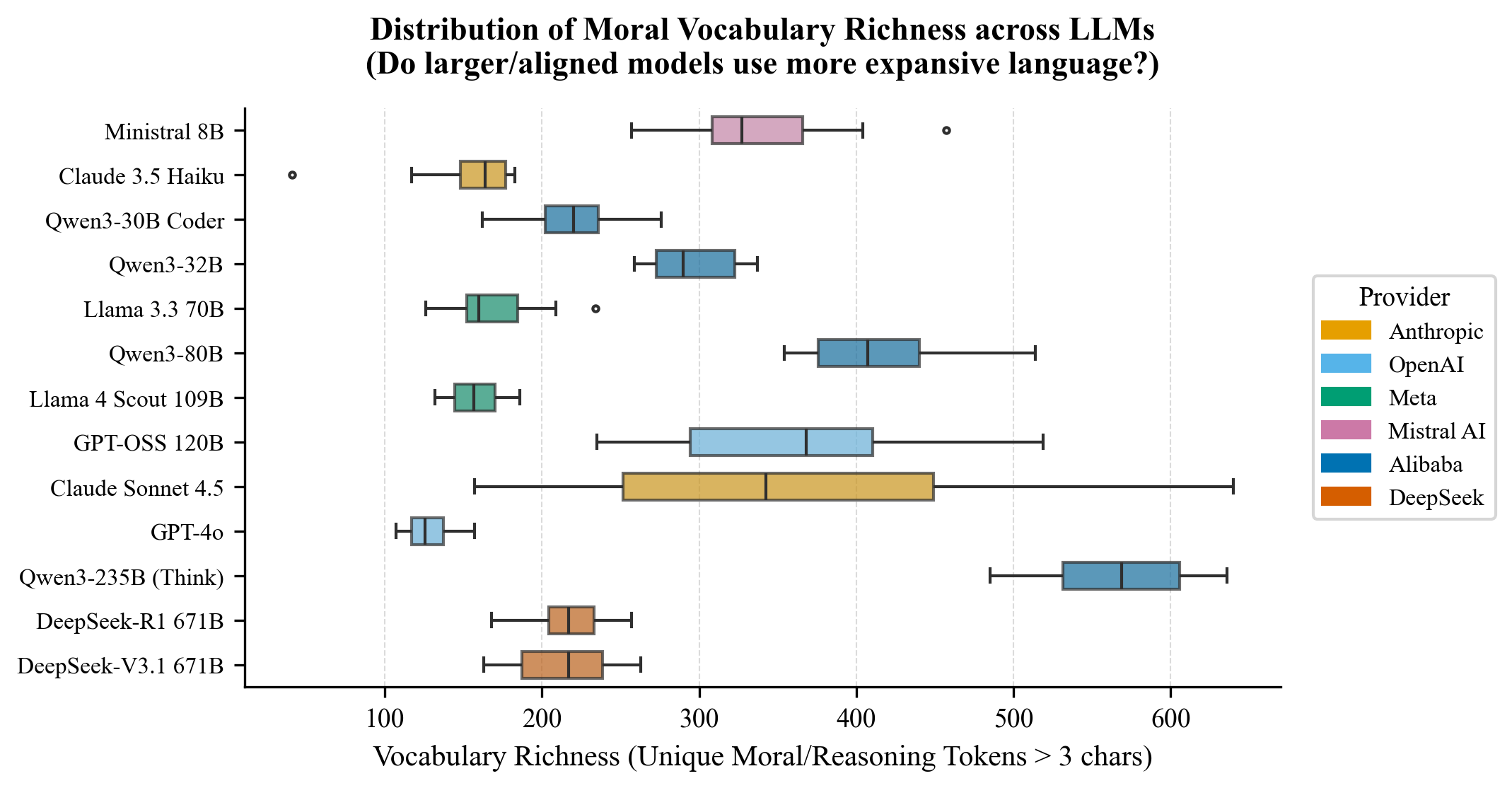}
  \caption{\textbf{RQ6: Distribution of Moral Vocabulary Richness across LLMs.} Horizontal box plots of the count of unique moral and reasoning tokens ($>$3 characters) per response for each model. Claude Sonnet~4.5 and Qwen3-235B~(Thinking) display the highest medians and widest spread, reflecting their tendency toward elaborate justifications. Llama models and GPT-4o cluster at the low end despite their large scale, suggesting that vocabulary richness is driven more by instruction-tuning style than raw parameter count.}
  \label{fig:vocab_richness}
\end{figure}

\begin{figure}[H]
  \centering
  \includegraphics[width=0.70\textwidth]{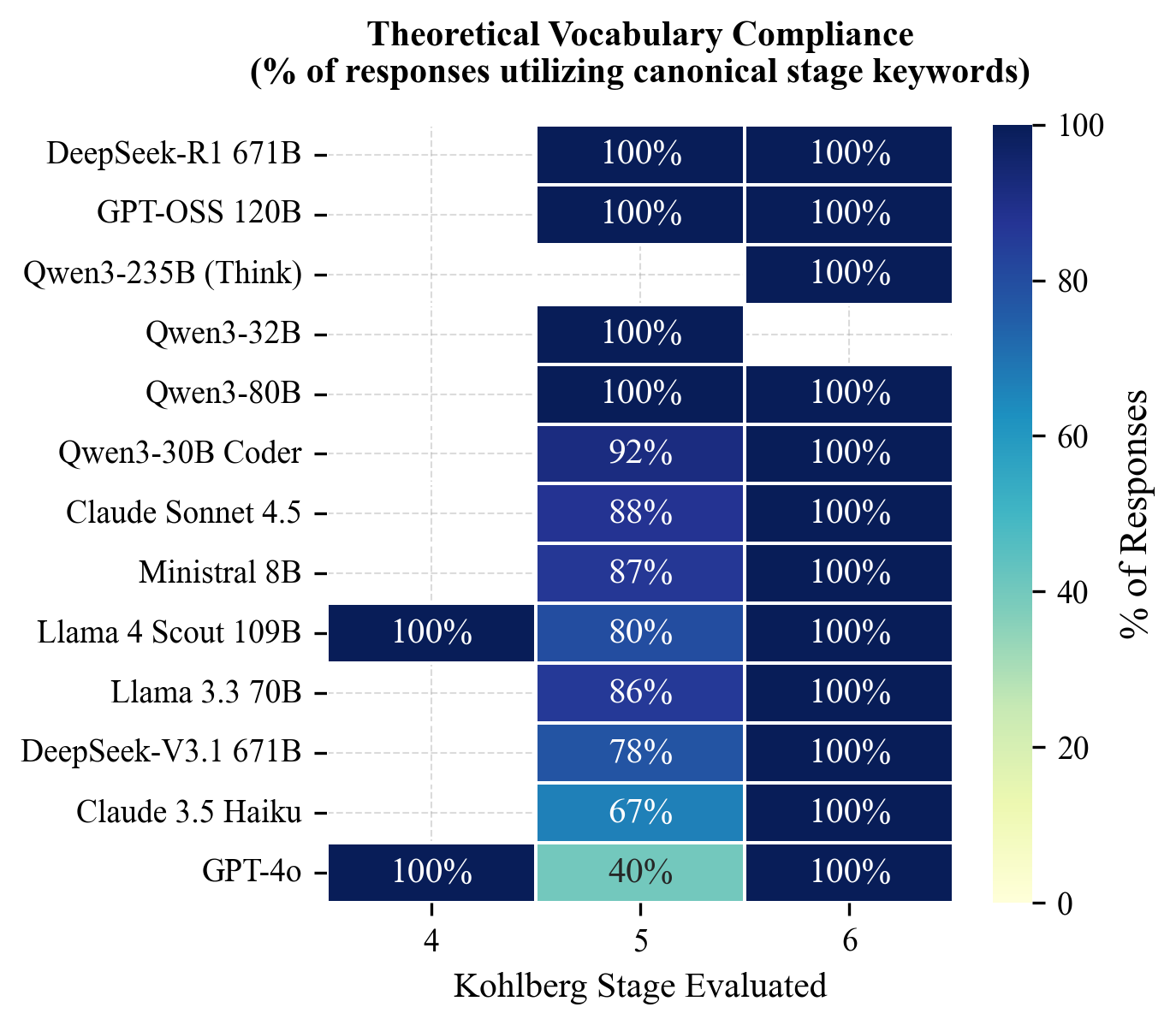}
  \caption{\textbf{RQ6: Theoretical Vocabulary Compliance Heatmap.} Percentage of responses utilizing canonical stage-specific keywords for each model at each Kohlberg stage it was evaluated on. Nearly all models achieve 100\% compliance at Stage~6, while Stage~5 compliance varies (40\%--100\%) with GPT-4o showing the largest shortfall. Stage~4 compliance applies only to Llama~4~Scout and GPT-4o, both of which occasionally produce Stage~4 evaluations, with perfect keyword usage. High compliance confirms that the evaluator scaffolding elicits stage-appropriate vocabulary even when broader reasoning patterns remain post-conventional.}
  \label{fig:keyword_heatmap}
\end{figure}

\FloatBarrier
\subsection{Sub-Capability Predictors of Moral Stage}
\label{app:subcap}

\begin{figure}[H]
  \centering
  \includegraphics[width=\textwidth]{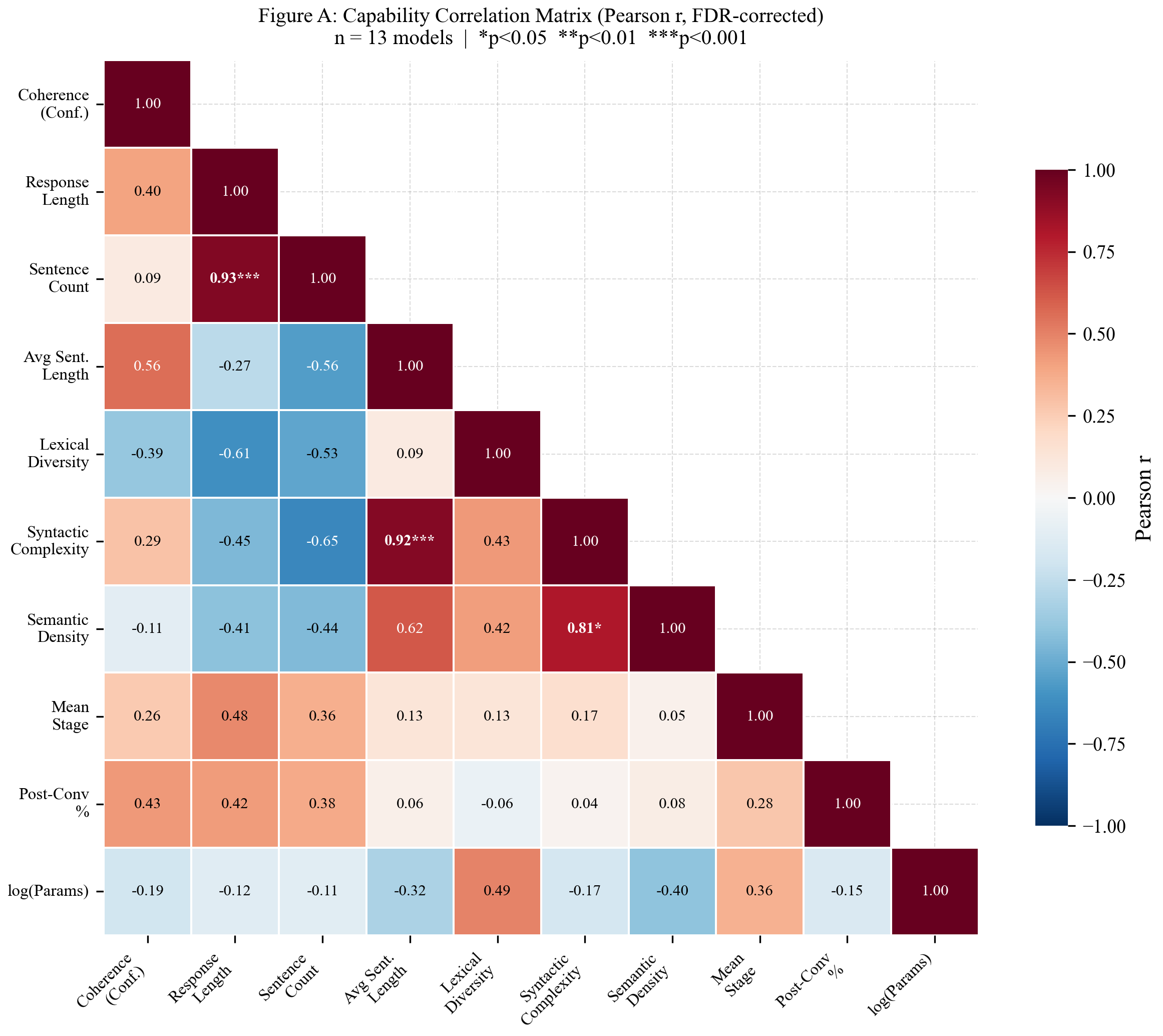}
  \caption{\textbf{Capability Correlation Matrix (Pearson $r$, FDR-corrected).} Pairwise correlations among eight response-level capability metrics and mean moral reasoning stage across 13 models. Response Length and Sentence Count are strongly correlated with each other ($r=0.93^{***}$), as are Average Sentence Length and Syntactic Complexity ($r=0.92^{***}$). Neither metric shows a significant correlation with Mean Stage or Post-Conventional percentage, confirming that surface linguistic elaboration does not predict moral sophistication in the Kohlberg sense.}
  \label{fig:cap_corr}
\end{figure}

\begin{figure}[H]
  \centering
  \includegraphics[width=0.80\textwidth]{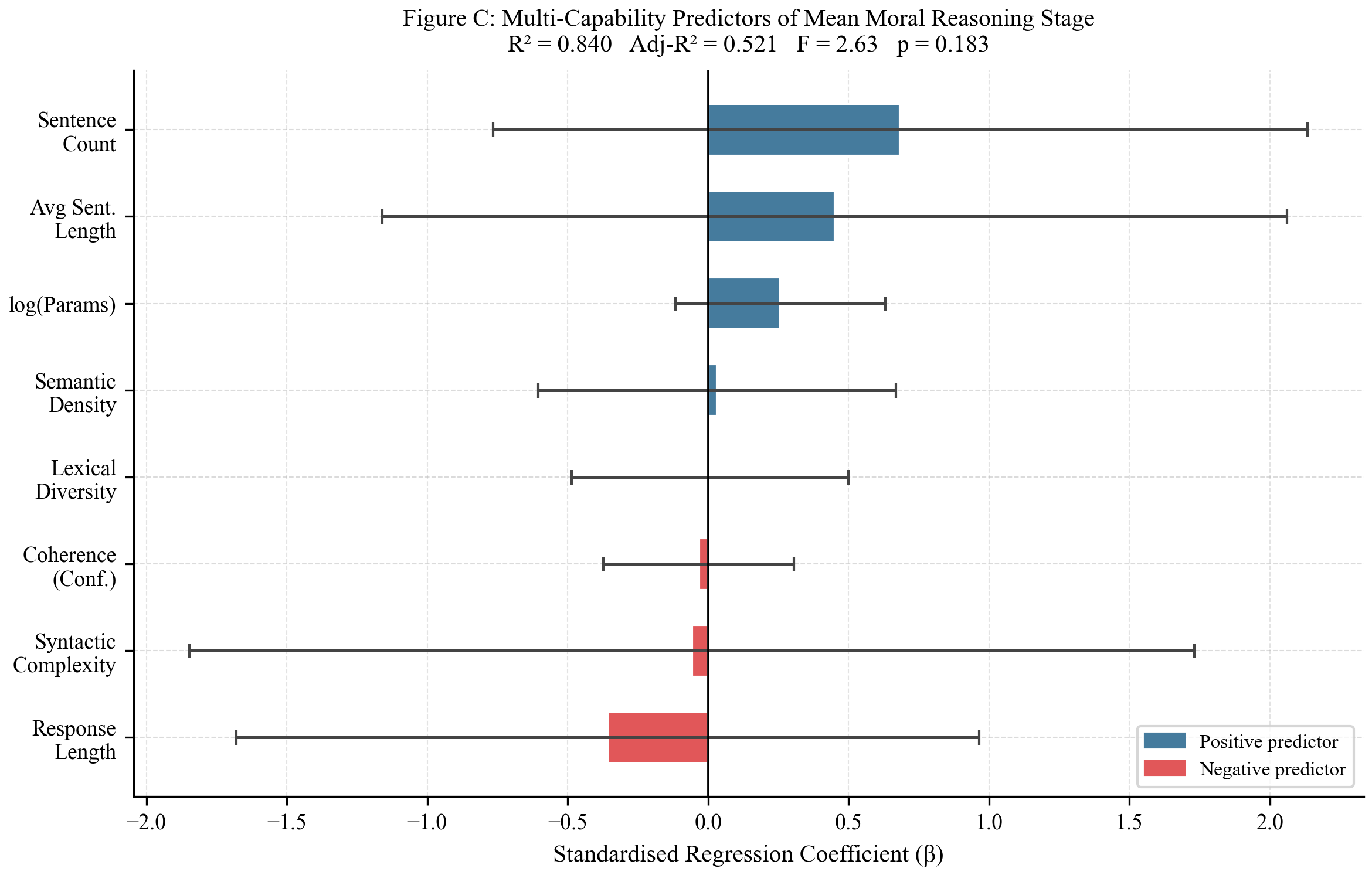}
  \caption{\textbf{Multi-Capability Predictors of Mean Moral Reasoning Stage (Standardized Coefficients).} Forest plot of standardized regression coefficients ($\beta$) from a multiple regression predicting mean stage from eight capability predictors ($R^2=0.840$, Adj-$R^2=0.521$, $p=0.183$). Sentence Count carries the largest positive coefficient, while Response Length is the strongest negative predictor, reflecting multicollinearity between elaboration metrics. Wide confidence intervals and a non-significant overall $F$-test indicate that no individual capability dimension reliably predicts moral stage, consistent with the broader finding that conventional measures of language capability do not map onto Kohlberg stage.}
  \label{fig:cap_regr}
\end{figure}

\FloatBarrier
\subsection{Stage Transition Dynamics}
\label{app:transitions}

\begin{figure}[H]
  \centering
  \includegraphics[width=0.55\textwidth]{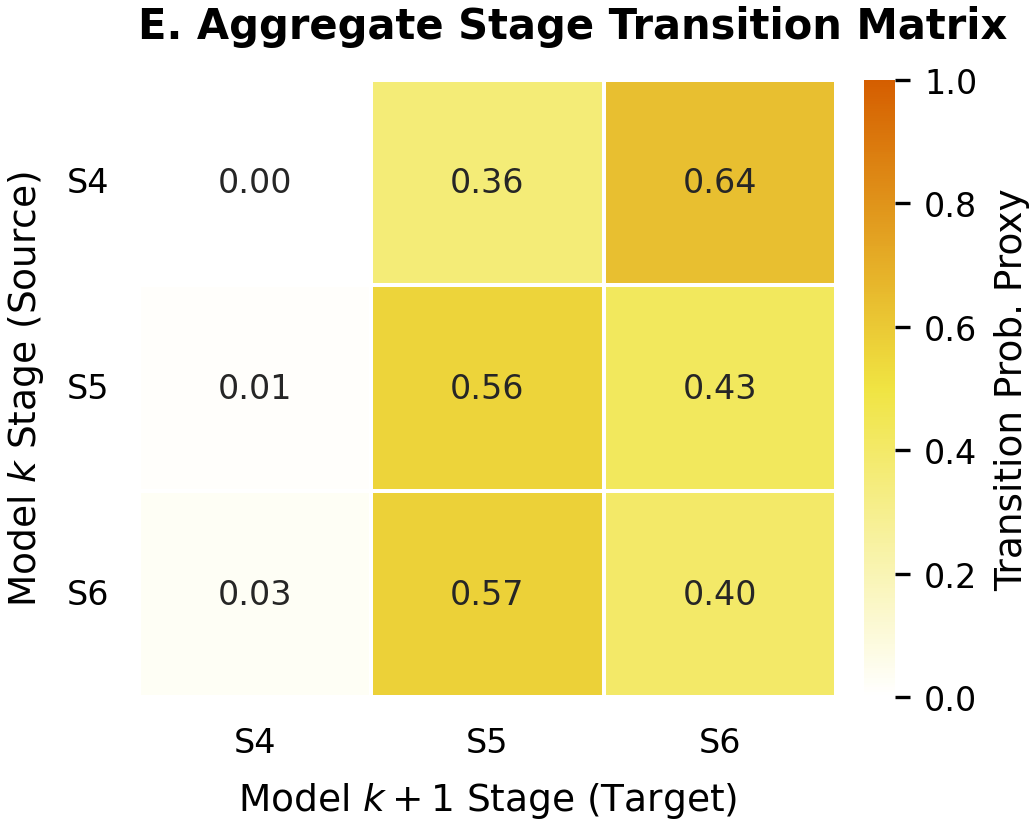}
  \caption{\textbf{Aggregate Stage Transition Matrix.} Heatmap of inter-model stage transition probabilities (rows = source model stage; columns = target model stage) when models are ordered by parameter count. Transitions from Stage~4 overwhelmingly target Stage~6 (0.64), suggesting a discontinuous jump rather than a gradual progression. Stage~5 models transition equally to Stage~5 (0.56) and Stage~6 (0.43), while Stage~6 models most often stay at Stage~5 in the next (smaller) model, implying mean-reversion in the scale-ordered sequence. These patterns are consistent with quantized rather than smooth scaling behavior in moral reasoning stage.}
  \label{fig:transition_matrix}
\end{figure}

\FloatBarrier
\subsection{Scale-Ordered Moral Stage Distribution}
\label{app:scale_dist}

\begin{figure}[H]
  \centering
  \includegraphics[width=0.80\textwidth]{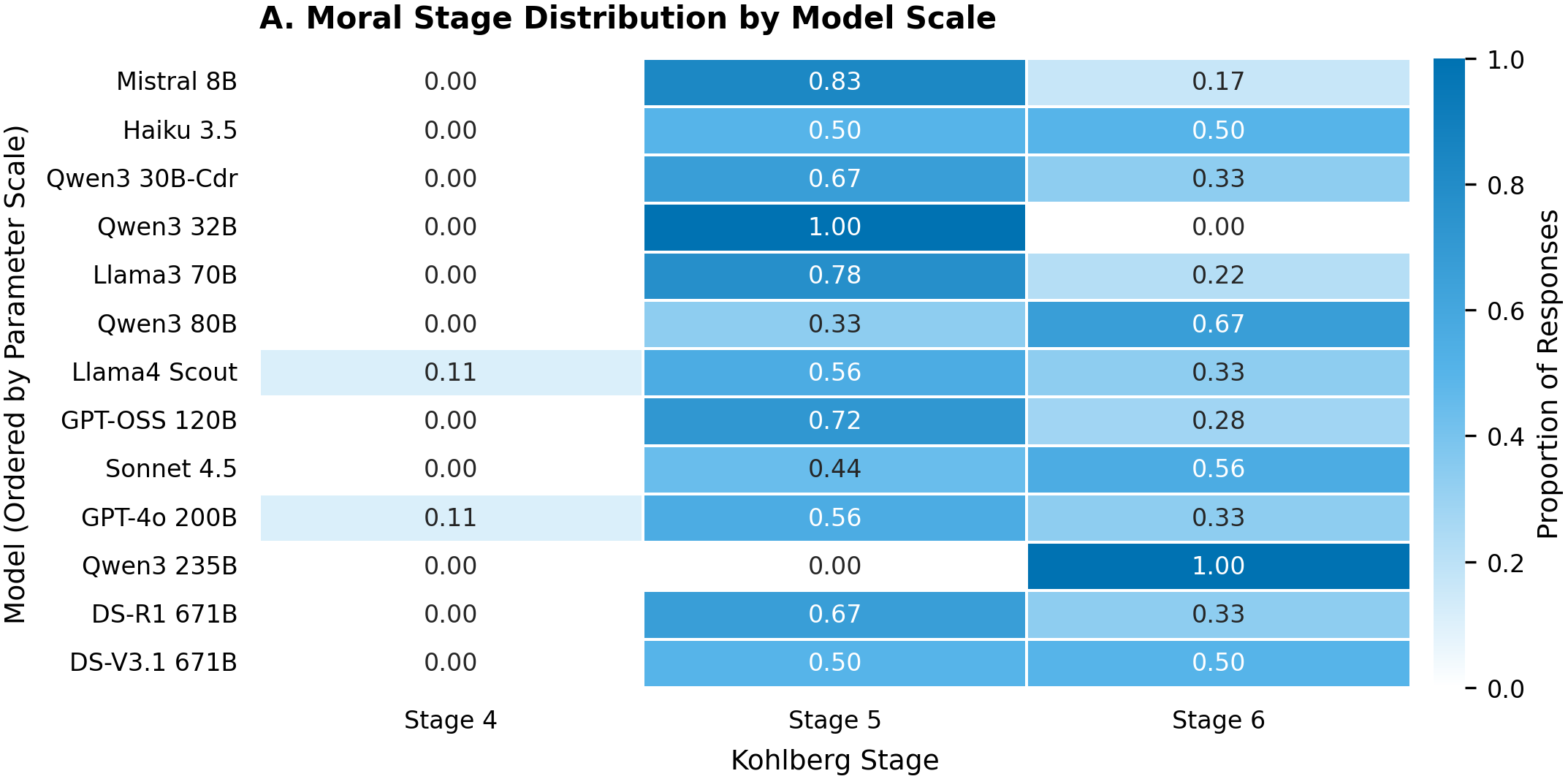}
  \caption{\textbf{Moral Stage Distribution by Model Scale.} Heatmap of the proportion of responses at each Kohlberg stage for 13 models ordered by increasing parameter count (bottom to top). Virtually all responses fall in Stage~5 or Stage~6; Stage~4 responses appear only for Llama~4~Scout and GPT-4o ($\approx$11\%). Qwen3-235B~(Thinking) reaches 100\% Stage~6, while Qwen3-32B achieves 100\% Stage~5 --- illustrating that the balance between these two post-conventional stages shifts as a function of scale and instruction-tuning orientation, yet neither model produces qualitatively different moral reasoning relative to the human developmental baseline.}
  \label{fig:scale_dist}
\end{figure}

\end{document}